\documentclass[runningheads]{llncs}
\usepackage{graphicx}

\usepackage{tikz}
\usepackage{comment}
\usepackage{amsmath,amssymb} 
\usepackage{booktabs}
\usepackage{color}
\usepackage{orcidlink}

\usepackage[accsupp]{axessibility}  

\usepackage[title]{appendix}
\usepackage{subcaption}
\usepackage{multirow}
\usepackage{bbm}
\usepackage{hyperref}
\hypersetup{
    bookmarks=false,
	colorlinks=true,
	linkcolor=blue,
	filecolor=blue,
    urlcolor=blue,
	citecolor=blue,
}

\begin{document}
\pagestyle{headings}
\mainmatter
\def\ECCVSubNumber{4929}  

\title{Few-Shot Classification with Contrastive Learning} 

\titlerunning{Few-Shot Classification with Contrastive Learning}
%
\author{Zhanyuan Yang\inst{1}\orcidlink{0000-0001-6449-7667} \and
Jinghua Wang\inst{2}\orcidlink{0000-0002-2629-1198} \and
Yingying Zhu\thanks{Corresponding author.}\inst{1}\orcidlink{0000-0002-3475-6186}}
\authorrunning{Yang et al.}
%
\institute{College of Computer Science and Software Engineering, Shenzhen University, Shenzhen, China\\
\email{yangzhanyuan2019@email.szu.edu.cn, zhuyy@szu.edu.cn}
\and
School of Computer Science and Technology, Harbin Institute of Technology (Shenzhen), Shenzhen, China\\
\email{wangjinghua@hit.edu.cn}}

\maketitle

\begin{abstract}
A two-stage training paradigm consisting of sequential pre-training and meta-training stages has been widely used in current few-shot learning (FSL) research.
Many of these methods use self-supervised learning and contrastive learning to achieve new state-of-the-art results.
However, the potential of contrastive learning in both stages of FSL training paradigm is still not fully exploited.
In this paper, we propose a novel contrastive learning-based framework that seamlessly integrates contrastive learning into both stages to improve the performance of few-shot classification.
In the pre-training stage, we propose a self-supervised contrastive loss in the forms of feature vector $vs.$ feature map and feature map $vs.$ feature map, which uses global and local information to learn good initial representations.
In the meta-training stage, we propose a cross-view episodic training mechanism to perform the nearest centroid classification on two different views of the same episode and adopt a distance-scaled contrastive loss based on them.
These two strategies force the model to overcome the bias between views and promote the transferability of representations.
Extensive experiments on three benchmark datasets demonstrate that our method achieves competitive results.
\end{abstract}

\section{Introduction}
\label{sec:intro}

Thanks to the availability of a large amount of annotated data, deep convolutional neural networks (CNN) \cite{ResNet,AlexNet,VGG} yield impressive results on various visual recognition tasks.
However, the time-consuming and costly collection process makes it a challenge for these deep learning-based methods to generalize in real-life scenarios with scarce annotated data.
Inspired by the capability of human to learn new concepts from a few examples, few-shot learning (FSL) is considered as a promising alternative to meet the challenge, as it can adapt knowledge learned from a few samples of base classes to novel tasks.

\begin{figure}
	\centering
    \includegraphics[width=0.39\columnwidth]{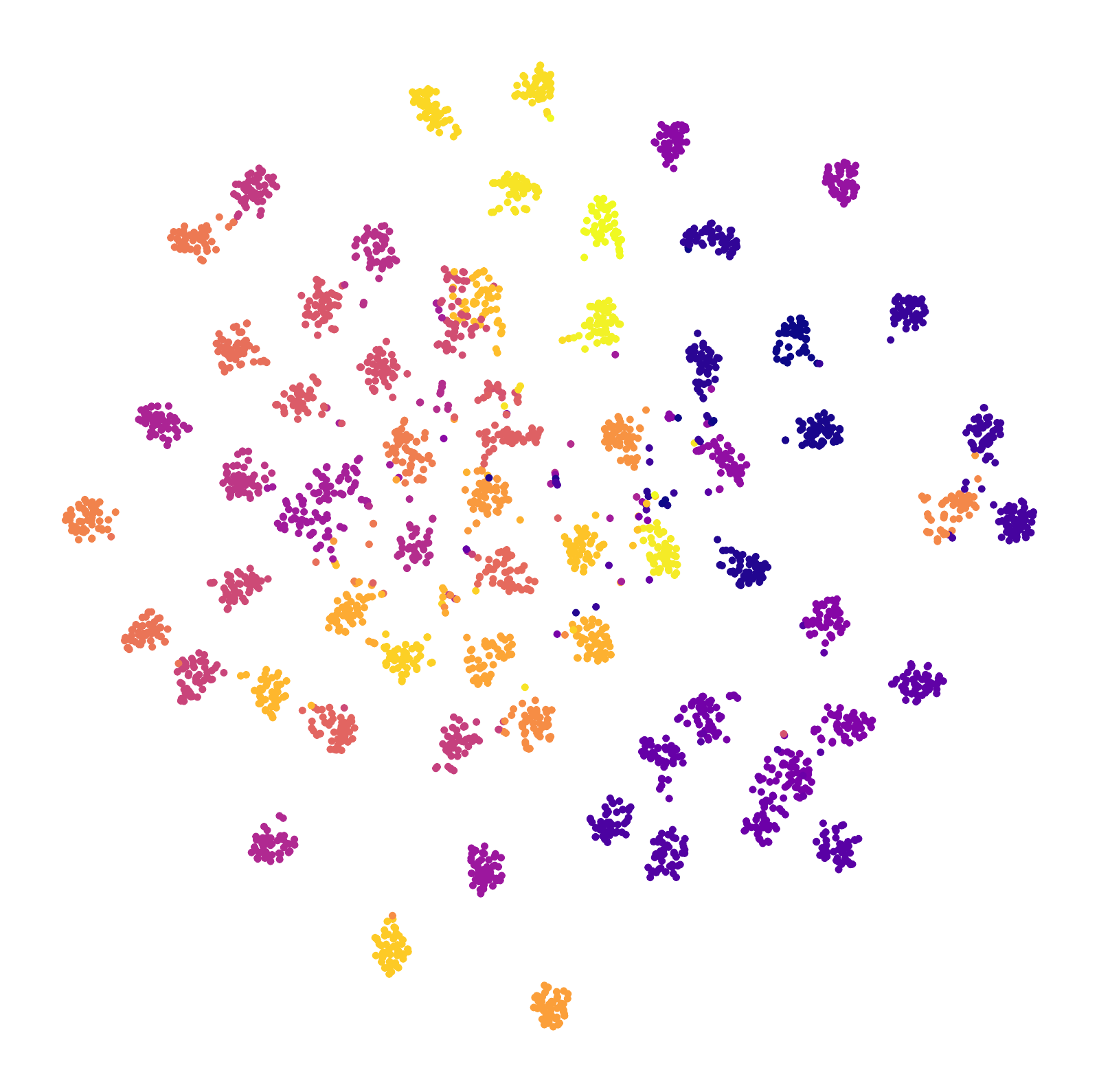}
    \includegraphics[width=0.39\columnwidth]{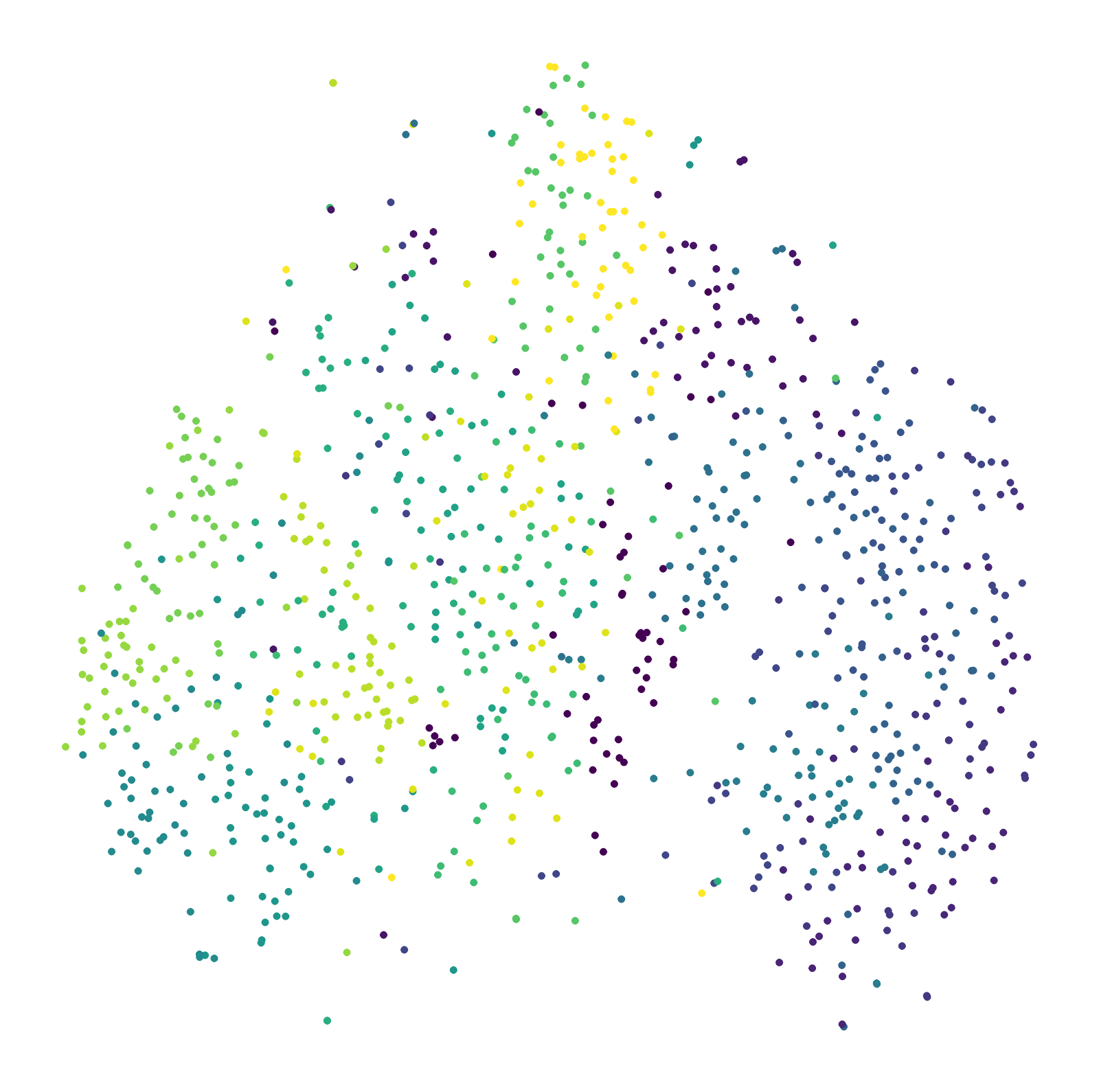}
	\caption{Distribution of feature embeddings of 64 base (\textit{left}) and 20 novel (\emph{right}) classes from miniImagenet in pre-train space by t-SNE~\cite{tsne}.}
	\label{fig:visual_pretrain}
\end{figure}
Recently, popular FSL methods \cite{MAML,Reptile,TADAM,LEO,ProtoNet,RelationNet,MatchNet} mainly adopt the meta-learning strategy.
These meta-learning based methods typically take episodic training mechanism to perform meta-training on base classes with abundant data.
During meta-training, the episodes consist of a support set and a query set, which are used in few-shot classification to mimic the evaluation setting.
The learned model is expected to be capable of generalizing across novel tasks of FSL.
Besides, many other methods~\cite{Closelook,InfoPatch,rfs,SimpleShot,TPMN,FEAT} achieve good classification accuracy by pre-training the feature extractor on base classes.
These methods suggest that the transferable and discriminative representations learned through pre-training or meta-training is crucial for few-shot classification.

However, both the pre-training and meta-training procedures only minimize the standard cross-entropy (CE) loss with labels from base classes.
The resulting models are optimized to solve the classification tasks of base classes.
Due to this, these methods may discard the information that might benefit the classification tasks on the unseen classes.
\figurename~\ref{fig:visual_pretrain} shows that the pre-trained model is able to identify samples from the base classes (left) well but performs poorly on samples from the novel classes (right).
That is, the learned representations are somewhat overfitted on the base classes and not generalizable on the novel classes.
Owing to the label-free nature of self-supervised learning methods, some recent works \cite{PSST,Boost,whendose} have tried self-supervised pretext tasks to solve the FSL problem, while other works \cite{CrossTransformers,InfoPatch,PAL,SCL} focus on contrastive learning methods.
Though promising, these approaches ignore the additional information from the self-supervised pretext tasks in meta-training or treat them as auxiliary losses simply in the FSL training paradigm.

In this work, we propose a contrastive learning-based framework that seamlessly integrates contrastive learning into the pre-training and meta-training stages to tackle the FSL problem.
First, in the pre-training stage,
we propose two types of contrastive losses based on self-supervised and supervised signals, respectively, to train the model.
These losses consider the global and local information simultaneously.
Our proposed self-supervised contrastive loss exploits local information in the forms of both feature vector $vs.$ feature map (vector-map) and feature map $vs.$ feature map (map-map), which differs from previous methods. 
Our supervised contrastive loss makes good use of the correlations among individual instances and the correlations among different instances of the same category.
Second, in the meta-training stage, motivated by the idea of maximizing mutual information between features extracted from multiple views (\emph{e.g.}, by applying different data augmentation on images) of the shared context (\emph{e.g.}, original images)~\cite{AMDIM,DIM,CPC,NPID}, we introduce a cross-view episodic training (CVET) mechanism to extract generalizable representations.
Concretely,
we randomly employ two different data augmentation strategies~\cite{SimCLR,MoCo,FEAT} to obtain the augmented episodes and treat them as different views of the original one.
Note, the augmentation does not change the label of the data.
We then conduct the nearest centroid classification between the augmented episodes to force the model to overcome the bias between views and generalize well to novel classes.
As a complement to CVET, we take inter-instance distance scaling into consideration and perform query instance discrimination within the augmented episodes.
These two methods effectively apply contrastive learning to the meta-training stage of FSL.
Our proposed method learns meta-knowledge that can play a crucial role in recognizing novel classes. 
The key contributions of this work are as follows:
\begin{itemize}
    \item We propose a contrastive learning-based FSL framework consisting of the pre-training and meta-training stages to improve the few-shot image classification. Our framework is easy to combine with other two-stage FSL methods. 
    \item We adopt the self-supervised contrastive loss based on global and local information in the pre-training stage to enhance the generalizability of the resulting representations.   
    \item We propose a CVET mechanism to force the model to find more transferable representations by executing classification between augmented episodes. 
    Meanwhile, we introduce a distance-scaled contrastive loss based on the augmented episodes to ensure that the classification procedure is not affected by extreme bias between different views. 
    \item Extensive experiments of few-shot classification on three benchmarks show that our proposed method achieves competitive results.
\end{itemize}


\section{Related Work}
\label{sec:relatedwork}

\subsection{Few-Shot Learning}
FSL aims to learn patterns on a large number of labeled examples called base classes and adapt to novel classes with limited examples per class.
Few-shot image classification has received great attention and many methods have been proposed.
The existing methods can be broadly divided into two categories: \emph{optimization-based} and \emph{metric-based}.
The \emph{optimization-based} methods initialize the model on base classes and adapt to novel tasks efficiently within a few gradient update steps on a few labeled samples \cite{MAML,MetaSGD,Reptile,LSTM-optimizer,LEO}.
The \emph{metric-based} methods aim to learn a generalizable representation space and use a well-defined metric to classify them \cite{DN4,TADAM,ProtoNet,RelationNet,MatchNet,AM3,DeepEMD}.
The existing works have considered different metrics such as cosine similarity \cite{MatchNet}, Euclidean distance \cite{ProtoNet}, a CNN-based relation module \cite{RelationNet}, a task-adaptive metric \cite{TADAM}, a local descriptor based metric \cite{DN4} and graph neural networks \cite{few-shotGNN}.
The Earth Mover's Distance \cite{DeepEMD} is employed as a metric to learn more discriminative structured representations.
Many recent studies \cite{Closelook,P-Transfer,rfs,SimpleShot} have proposed a standard end-to-end pre-training framework to obtain feature extractors or classifiers on base classes.
These pre-training based methods achieve competitive performance compared to episodic meta-training methods.
Moreover, many papers \cite{meta-baseline,InfoPatch,FEAT,DeepEMD} take advantage of a sequential combination of pre-training and meta-training stages to further enhance the performance.
The methods \cite{CAN,RelationNet,TPMN,DMF,FEAT} pay more attention to the transferability of representations through delicately designing task-specific modules in meta-training.
Given the simplicity and effectiveness of these methods, we take FEAT \cite{FEAT} as our baseline, but drop its auxiliary loss.

\subsection{Contrastive Learning}
Recently, contrastive learning with the instance discrimination as a pretext task has become a dominant approach in self-supervised representation learning \cite{AMDIM,SimCLR,MoCo,DIM,CPC,CMC,NPID}.
These methods typically construct contrast pairs of instances with a variety of data augmentation and optimize a contrastive loss with the aim of keeping instances close to their
augmented counterparts while staying away from other instances in the embedding space.
The goal of contrastive learning using self-supervision from instances is to improve the generalizability of the representations and benefit various downstream tasks.
Contrastive learning is also extended to group instances in a supervised manner~\cite{SupContrast} and achieves better performance than CE loss on standard classification tasks.

\subsection{Few-Shot Learning with Contrastive Learning}
In contrast to the works \cite{PSST,Boost,whendose} that introduce self-supervised pretext tasks such as rotation \cite{Rotation} and jigsaw \cite{Jigsaw} into FSL as auxiliary losses,
recent approaches \cite{CrossTransformers,InfoPatch,PAL,SCL} have explored contrastive learning of instance discrimination in different parts of the two-stage training pipeline of FSL.
Methods \cite{InfoPatch,PAL,SCL} combine supervised contrastive loss \cite{SupContrast} to the pre-training stage \cite{PAL,SCL} and the meta-training stage \cite{CrossTransformers,InfoPatch}, respectively.
Unlike prior works, our proposed method boosts few-shot classification performance by seamlessly integrating instance-discriminative contrastive learning in both the pre-training and meta-training stages.
In the pre-training stage, we conduct self-supervised contrastive loss in the forms of vector-map and map-map.
In the meta-training stage, we combine contrastive learning with episodic training and define a distance-scaled contrastive loss to improve the transferability of the representations.
%

\section{Method}
\label{sec:method}
\begin{figure}[t]
    \centering
    \includegraphics[width=1.\columnwidth]{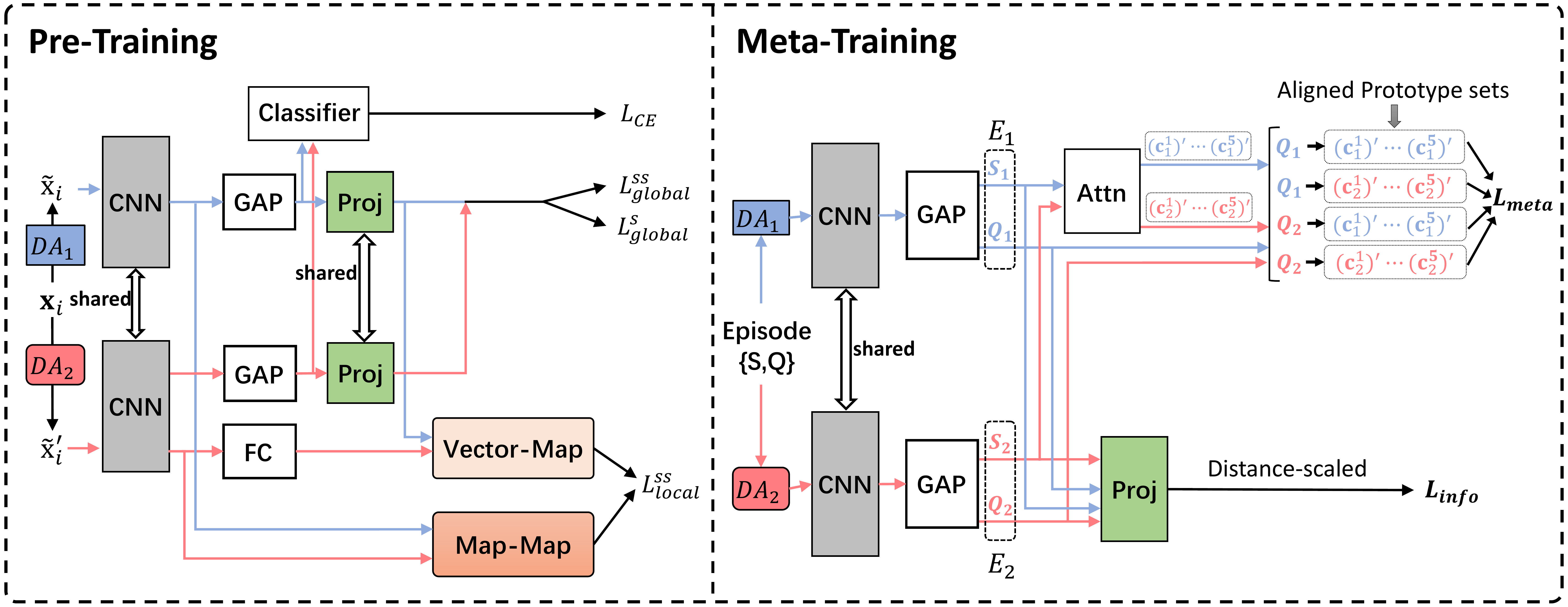}
    \caption{Overview of our framework.
    Based on multiple views of an input through two random data augmentation $DA_1$ and $DA_2$, we compute contrastive losses at both global and local levels in the pre-training stage.
    In the meta-training stage, we enforce cross-view episodic training and compute a distance-scaled contrastive loss episodically.
    Here, GAP dentoes a global average pooling layer, Proj is a projection head, FC means a fully contected layer, Attn is the task-specific module from \cite{FEAT}, $\mathcal{S}_{r},\mathcal{Q}_{r}$ mean support and query set from different views of the episode $E=\{\mathcal{S}, \mathcal{Q}\}$ respectively, and $(c_{r}^k)^{\prime}$ dentoes the aligned prototype in $S_{r}$.
    }
    \label{fig:overview}
\end{figure}

\subsection{Preliminary}
The few-shot classification task is slightly different from the standard supervised classification task.
The meta-training set $\mathcal{D}_{train}=\left\{\left(x_{i}, y_{i}\right) \mid y_{i} \in \mathcal{C}_{base}\right\}$ consists of the samples from the base classes $\mathcal{C}_{base}$ and the meta-test set $\mathcal{D}_{test}=\left\{\left(x_{i}, y_{i}\right) \mid y_{i} \in \mathcal{C}_{novel}\right\}$ consists of the samples from the novel classes $\mathcal{C}_{novel}$.
Here, $y_{i}$ is the class label of sample $x_{i}$.
In FSL, we aim to learn a model based on $\mathcal{D}_{train}$ and generalize it over $\mathcal{D}_{test}$,
where $\mathcal{C}_{base} \cap \mathcal{C}_{novel} = \emptyset$.
Following the prior meta-learning based methods \cite{MAML,ProtoNet,MatchNet}, we adopt episodic mechanism to simulate the evaluation setting.
Concretely, each $M-$way $K-$shot episode $E$ consists of a support set and a query set.
We first randomly sample $M$ classes from $\mathcal{C}_{base}$ for meta-training (or from $\mathcal{C}_{novel}$ for meta-testing) and $K$ instances per class to obtain the support
set $\mathcal{S}=\left\{x_{i}, y_{i}\right\}_{i=1}^{M*K}$.
Then, we sample $Q$ instances in each of the selected classes to obtain the query set $\mathcal{Q}=\left\{x_{i}, y_{i}\right\}_{i=1}^{M*Q}$.
Note that $y_{i} \in \left\{1,2,\ldots,M\right\}$ and $\mathcal{S} \cap \mathcal{Q} = \emptyset$.
The episodic training procedure classifies the samples in $\mathcal{Q}$ into the categories corresponding to the samples in $\mathcal{S}$.

\subsection{Overview}
In this work, we follow the two-stage training strategy and incorporate contrastive learning in both stages to learn more generalizable representations.
Our proposed framework is illustrated in \figurename~\ref{fig:overview}.
In the pre-training stage, we adopt self-supervised and supervised contrastive losses to obtain a good initial representation.
In the meta-training stage, we propose a novel cross-view episodic training (CVET) mechanism and a distance-scaled contrastive loss,
which allows the model to overcome the bias between views of each episode and generalize well across novel tasks.
Note that we take FEAT \cite{FEAT} without its auxiliary loss as the baseline due to its simple and effective task-specific module (a multi-head attention module~\cite{AttentionIsAllYouNeed}).
We will detail our framework in the following subsections.

\subsection{Pre-training}
\label{sec:pretrain}
In this section, we introduce instance-discriminative contrastive learning \cite{SimCLR,SupContrast} in the pre-training stage to alleviate the overfitting problem caused by training with CE loss only.
As shown in \figurename~\ref{fig:overview}, we propose self-supervised contrastive losses at the global and local levels, respectively.
Using self-supervision in these losses helps produce more generalizable representations.
Meanwhile, we also employ a global supervised contrastive loss \cite{SupContrast} to capture the correlations among instances from the same category.

\noindent \textbf{Global self-supervised contrastive loss.}
This loss (\emph{a.k.a} InfoNCE loss~\cite{SimCLR,CPC}) aims to enhance the similarity between the views of the same image, while reducing the similarity between the views of different images.
Formally, we randomly apply two data augmentation methods to a batch of samples $\left\{x_i, y_i\right\}^{N}_{i=1}$ from the meta-training set $\mathcal{D}_{train}$ and generate the augmented batch $\left\{\widetilde{x}_i, \widetilde{y}_i\right\}^{2N}_{i=1}$.
Here, $\widetilde{x}_{i}$ and $\widetilde{x}_{i}^{\prime}$ denote two different views of $x_i$,  which are considered as a positive pair.
We define $f_{\phi}$ as the feature extractor with learnable parameters $\phi$ to transform the sample $\widetilde{x}_i$ into a feature map $\mathbf{\hat{x}}_i = f_{\phi}(\widetilde{x}_i)\in \mathbb{R}^{C \times H \times W}$ and further obtain the global feature $\mathbf{h}_{i} \in \mathbb{R}^{C}$ after a global average pooling (GAP) layer.
We use a MLP with one hidden layer to instantiate a projection head $proj(\cdot)$ \cite{SimCLR} to generate the projected vector $\mathbf{z}_{i}=proj(\mathbf{h}_{i}) \in \mathbb{R}^{D}$.
Then the global self-supervised contrastive loss can be computed as:
\begin{equation}
	\footnotesize
    {L}^{ss}_{global}=-\sum_{i=1}^{2N} \log \frac{\exp \left(\mathbf{z}_{i} \cdot \mathbf{z}_{i}^{\prime} / \tau_{1}\right)}{\sum_{j=1}^{2N} \mathbbm{1}_{j \neq i}\exp \left(\mathbf{z}_{i} \cdot \mathbf{z}_{j} / \tau_{1}\right)},
    \label{eq:ss_global}
\end{equation}
where the $\cdot$ operation denotes inner product after $l_2$ normalization, $\tau_{1}$ is a scalar temperature parameter, and $\mathbbm{1} \in \left\{0,1\right\}$ is an indicator function.
Here, the positive pair, $\mathbf{z}_{i}^{\prime} $ and $\mathbf{z}_{i}$, are extracted from the augmented versions of the same sample $x_{i}$.

\begin{figure}[!t]
\centering
    \begin{minipage}[t]{0.49\columnwidth}
        \includegraphics[width=1.0\linewidth]{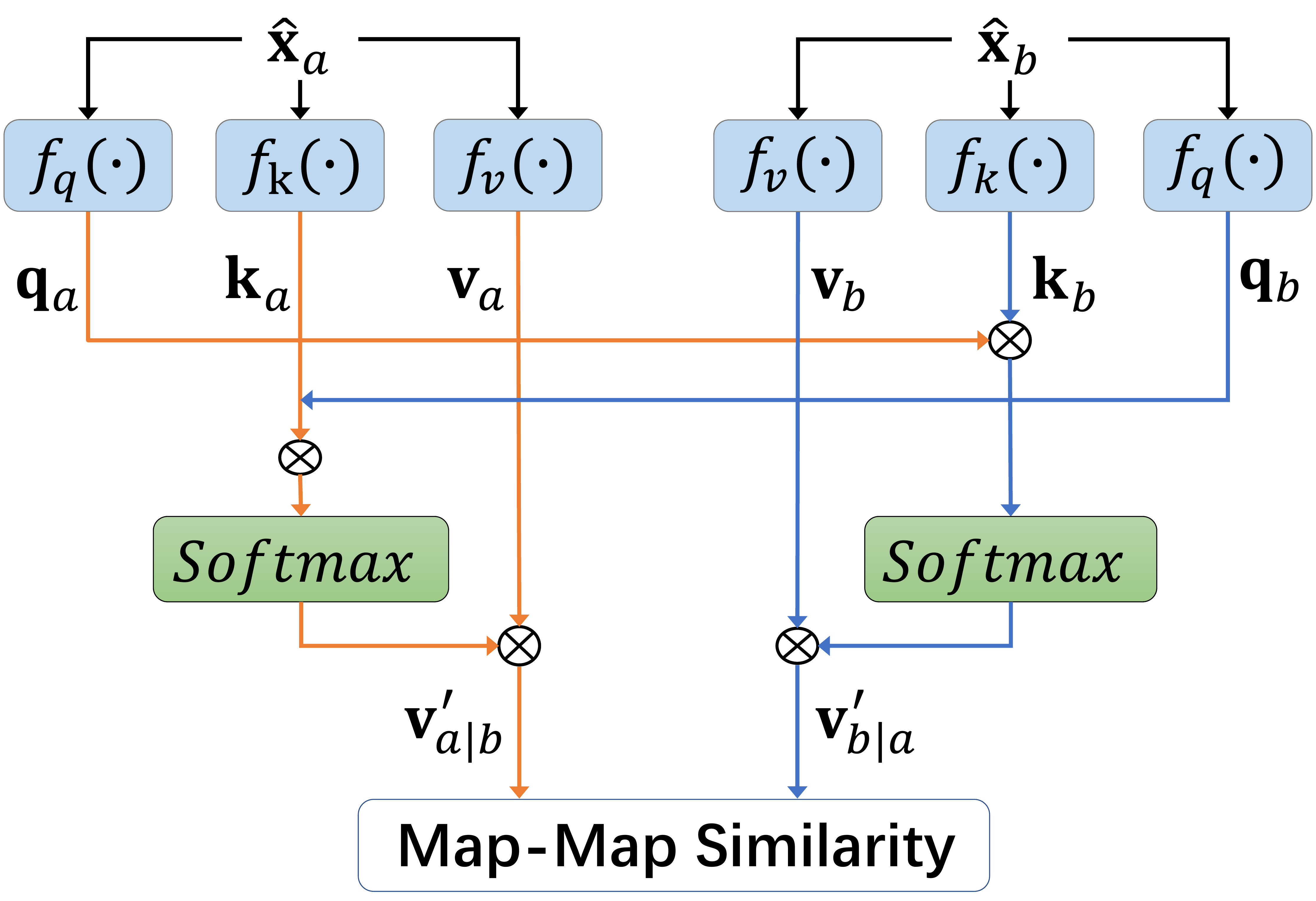}
        \subcaption{Map-map module}
        \label{fig:map_to_map}
    \end{minipage}
    \begin{minipage}[t]{0.49\columnwidth}
        \includegraphics[width=1.0\linewidth]{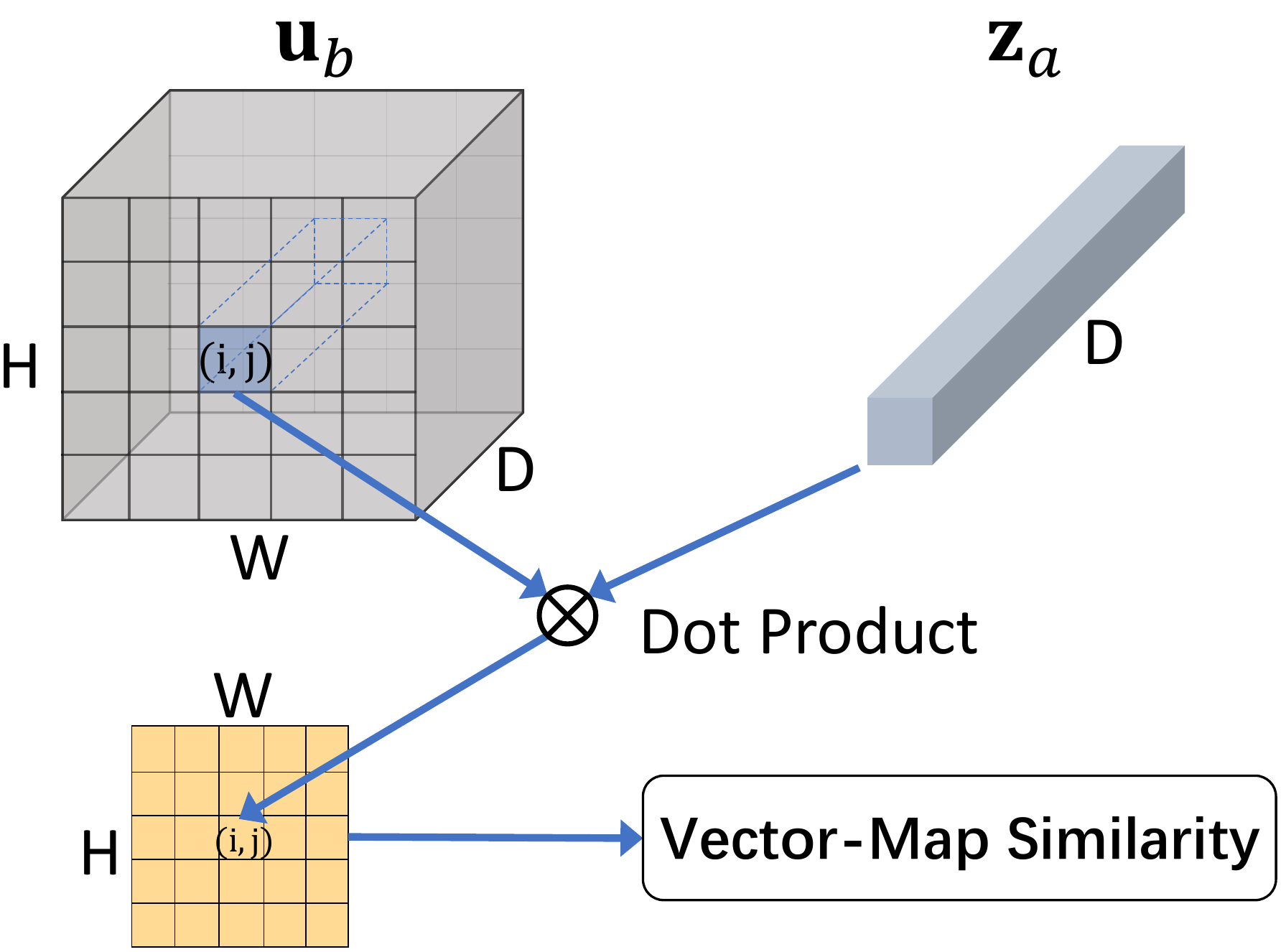}
        \subcaption{Vector-map module}
        \label{fig:vector_to_map}
    \end{minipage}
    \caption{Map-map and vector-map modules. (a) Operation $\otimes$ denotes dot product. Local feature maps $\hat{\mathbf{x}}_a$ and $\hat{\mathbf{x}}_b$ share three spatial projection heads $f_{q}(\cdot)$, $f_{k}(\cdot)$ and $f_{v}(\cdot)$. We first align $\hat{\mathbf{x}}_a$ with $\mathbf{q}_b$ and then align $\hat{\mathbf{x}}_b$ with $\mathbf{q}_a$.
    (b) We obtain $\mathbf{u}_{b}$ by adding a FC layer after $\hat{\mathbf{x}}_b$ and $\mathbf{z}_a$ denotes the projected vector from $proj(\hat{\mathbf{x}}_a)$.}
\end{figure}
\noindent \textbf{Local self-supervised contrastive loss.}
Though ${L}^{ss}_{global}$ (Eq.~\eqref{eq:ss_global}) favors transferable representations based on global feature vector $\mathbf{h}_i$, it might ignore some local discriminative information in feature map $\hat{\mathbf{x}}_i$ which could be beneficial in meta-testing.
Inspired by \cite{AMDIM,SimCLR,DIM,SCL},
we compute self-supervised contrastive loss at the local level.
Unlike previous approaches, we leverage map-map and vector-map modules to boost the robustness and generalizability of the representations.
The \textbf{map-map module} is illustrated in \figurename~\ref{fig:map_to_map}.
Specifically, we use three $f_{q},f_{k},f_{v}$ spatial projection heads to project local feature map $\hat{\mathbf{x}}_i$ into
the query $\mathbf{q}_i=f_{q}(\hat{\mathbf{x}}_i)$, key $\mathbf{k}_i=f_{k}(\hat{\mathbf{x}}_i)$ and value $\mathbf{v}_i=f_{v}(\hat{\mathbf{x}}_i)$, respectively,
where $\mathbf{q}_i, \mathbf{k}_i, \mathbf{v}_i \in \mathbb{R}^{HW\times D}$.
For a pair of local feature map $\hat{\mathbf{x}}_a$ and $\hat{\mathbf{x}}_b$, we align the $\hat{\mathbf{x}}_a$ with $\hat{\mathbf{x}}_b$ to obtain $\mathbf{v}_{a\mid b}^{\prime} = softmax\left(\frac{\mathbf{q}_{b} \mathbf{k}^{\top}_{a}}{\sqrt{d}}\right) \mathbf{v}_{a}$, and align $\hat{\mathbf{x}}_b$ with $\hat{\mathbf{x}}_a$ to obtain $\mathbf{v}_{b\mid a}^{\prime} = softmax\left(\frac{\mathbf{q}_{a} \mathbf{k}^{\top}_{b}}{\sqrt{d}}\right) \mathbf{v}_{b}$.
After $l_2$ normalization on each position $(i,j)$ of the aligned results, we can compute the similarity between the two local feature maps $\hat{\mathbf{x}}_a$ and $\hat{\mathbf{x}}_b$ as follows:
\begin{equation}
\footnotesize
    sim_{1}\left(\hat{\mathbf{x}}_a, \hat{\mathbf{x}}_b\right) = \frac{1}{H W} \sum_{1\leq i\leq H,1\leq j\leq W}\left(\mathbf{v}_{a\mid b}^{\prime}\right)^{\top}_{ij} \left(\mathbf{v}_{b\mid a}^{\prime}\right)_{ij}.
\label{eq:spatial-similarity}
\end{equation}
Basically, Eq.~\eqref{eq:spatial-similarity} calculates the summation of the element-wise product of two feature maps $\mathbf{v}_{a\mid b}^{\prime}$ and $\mathbf{v}_{b\mid a}^{\prime} \in \mathbb{R}^{HW\times D}$.
The self-supervised contrastive loss based on pairwise feature maps can be computed as follows:
\begin{equation}
\footnotesize
    {L}^{ss}_{map-map}=-\sum_{i=1}^{2N} \log \frac{\exp \left(sim_{1}\left(\hat{\mathbf{x}}_{i}, \hat{\mathbf{x}}_{i}^{\prime}\right) / \tau_{2}\right)}{\sum_{j=1}^{2N} \mathbbm{1}_{j \neq i}\exp \left(sim_{1}\left(\hat{\mathbf{x}}_{i}, \hat{\mathbf{x}}_{j}\right) / \tau_{2}\right)},
    \label{eq:map_to_map}
\end{equation}
where $(\hat{\mathbf{x}}_{i}, \hat{\mathbf{x}}_{i}^{\prime})$ is a positive pair and ${\tau}_{2}$ denotes a temperature parameter, and $\mathbbm{1}$ is an indicator function.
Meanwhile, we adopt \textbf{vector-map module} to further exploit the local contrastive information between instances, which is shown in \figurename~\ref{fig:vector_to_map}.
In specific, we use a fully connected (FC) layer to obtain $\mathbf{u}_i = g(\hat{\mathbf{x}}_i)=\sigma(\mathbf{W}\hat{\mathbf{x}}_i) \in \mathbb{R}^{D\times HW}$, where $\sigma$ is a ReLU nonlinearity.
We can compute the similarity between a contrast pair as
    $sim_{2}\left(\hat{\mathbf{x}}_a, \hat{\mathbf{x}}_b\right)=\frac{1}{H W} \sum_{1\leq i\leq H,1\leq j\leq W}(\mathbf{u}_{b})_{ij}^{\top} \mathbf{z}_{a}$,
where $\mathbf{z}_{a}$ is the projected vector of $\hat{\mathbf{x}}_{a}$.
The self-supervised contrastive loss based on pairs of feature vectors and feature maps can be computed as follows:
\begin{equation}
    \footnotesize
    {L}^{ss}_{vec-map}=-\sum_{i=1}^{2N} \log \frac{\exp \left(sim_{2}\left(\hat{\mathbf{x}}_{i}, \hat{\mathbf{x}}_{i}^{\prime}\right) / \tau_{3}\right)}{\sum_{j=1}^{2N} \mathbbm{1}_{j \neq i}\exp \left(sim_{2}\left(\hat{\mathbf{x}}_{i}, \hat{\mathbf{x}}_{j}\right) / \tau_{3}\right)},
\end{equation}
where $\tau_{3}$ and $\mathbbm{1}$ act the same as in Eq.~\eqref{eq:map_to_map}.
Therefore, the local self-supervised contrastive loss can be defined as:
\begin{equation}
    \footnotesize
    {L}^{ss}_{local}={L}^{ss}_{vec-map} + {L}^{ss}_{map-map}.
\end{equation}

\noindent \textbf{Global supervised contrastive loss.}
To exploit the correlations among individual instances and the correlations among different instances from the same category,
we also adopt supervised contrastive loss~\cite{SupContrast} as follows:
\begin{equation}
    \footnotesize
    {L}^{s}_{global}=\sum_{i=1}^{2N} \frac{1}{|P(i)|}\sum_{p\in P(i)} L_{ip},
\end{equation}
where $L_{ip}=- \log \frac{\exp \left(\mathbf{z}_{i} \cdot \mathbf{z}_{p} / \tau_{4}\right)}{\sum_{j=1}^{2N} \mathbbm{1}_{j \neq i}\exp \left(\mathbf{z}_{i} \cdot \mathbf{z}_{j} / \tau_{4}\right)}$, and $\tau_{4}$ is a temperature parameter.
The set $P(i)$ contains the indexes of samples with the same label as $x_{i}$ in the augmented batch except for index $i$.

To summarize, we minimize the following loss during pre-training:
\begin{equation}
    \footnotesize
    L_{pre} = L_{CE} + \alpha_1 L^{ss}_{global} + \alpha_2 L^{ss}_{local} + \alpha_3 L^{s}_{global},
\end{equation}
where $L_{CE}$ is the CE loss, and $\alpha_1, \alpha_2$ and $\alpha_3$ are balance scalars.
By optimizing $L_{pre}$, we promote the discriminability and generalizability of the representations, which is crucial for the following stage.

\subsection{Meta-training}
\label{sec:crossview}
\textbf{Cross-view Episodic Training.}
In order to capture information about the high-level concept of a shared context,
a common practice in contrastive learning~\cite{SimCLR,MoCo,CPC,NPID} is to maximize the mutual information between features extracted from multiple views of the shared context.
Intuitively, given an episode $E=\left\{\mathcal{S}, \mathcal{Q}\right\}$ in episodic meta-learning, we can obtain two episodes $E_{1}=\left\{\mathcal{S}_{1}, \mathcal{Q}_{1}\right\}$ and $E_{2}=\left\{\mathcal{S}_{2}, \mathcal{Q}_{2}\right\}$ by applying two different data
augmentation strategies on $E$ respectively.
Here, we consider $E$ a shared context and treat the two augmented episodes as its two views.
Inspired by the above idea,
we propose a cross-view episodic training mechanism with the aim of forcing representations
to learn meta-knowledge that can play a key role across various few-shot classification tasks.
Specifically, we use the pre-trained feature extractor $f_{\phi}$ in Sect.~\ref{sec:pretrain} followed by a GAP to each data point $x_i$ in both the episode $E_{1}$ and $E_{2}$, and derive the corresponding global vector $\mathbf{h}_i \in \mathbb{R}^{C}$.
We then separately compute a prototype $\mathbf{c}_{r}^{k}$ for category $k$ in support sets $\mathcal{S}_1$ and $\mathcal{S}_2$ as follows:
\begin{equation}
\label{eq:meta_prototype}
\footnotesize
    \mathbf{c}_{r}^{k} = \frac{1}{|\mathcal{S}_{r}^{k}|} \sum_{(\mathbf{h}_i,y_i)\in \mathcal{S}_{r}^{k}} \mathbf{h}_{i},
\end{equation}
where $r\in \left\{1,2\right\}$ and $\mathcal{S}_{r}^{k}$ denotes the set of data points belonging to class $k\in\left\{1,2,\ldots,M\right\}$ from $r-{th}$ support set.
We denote the task-specific module (multi-head attention~\cite{AttentionIsAllYouNeed}) proposed in baseline~\cite{FEAT} as $Attn\left(\cdot\right)$ and fix the number of heads to 1.
With the task-specific module, we obtain an aligned prototype set $\mathcal{T}(r)=\{(\mathbf{c}_{r}^{k})^{\prime}\}_{k=1}^{M}=Attn\left(\mathbf{c}_{r}^{1},\mathbf{c}_{r}^{2}, \ldots,\mathbf{c}_{r}^{M}\right)$ for each support set.
Then, based on the aligned prototypes, a probability distribution of a data point $x_{i}$ in the query set $\mathcal{Q}_r$ over $M$ classes is defined as:
\begin{equation}
\footnotesize
    P(y=k \mid \mathbf{h}_{i}, {\mathcal{T}(r)})=\frac{\exp \left(-d\left(\mathbf{h}_i, (\mathbf{c}_{r}^{k})^{\prime}\right)\right)}{\sum_{j=1}^{M} \exp \left(-d\left(\mathbf{h}_i, (\mathbf{c}_{r}^{j})^{\prime}\right)\right)},
\end{equation}
where $d\left(\cdot,\cdot \right)$ denotes Euclidean distance.
Therefore the loss of the nearest centroid classifier on an episode can be computed as:
\begin{equation}
\footnotesize
    L_{mn}=\frac{1}{\left|\mathcal{Q}_{m}\right|} \sum_{\left(\mathbf{h}_{i}, y_{i}\right) \in \mathcal{Q}_{m}}-\log P(y=k \mid \mathbf{h}_{i}, \mathcal{T}(n)),
\label{eq:ncm}
\end{equation}
where $m, n \in \{1,2\}$.
Obviously different from the original episodic training, we classify $\mathcal{Q}_{1}$ on $\mathcal{S}_{1}$ and $\mathcal{S}_{2}$ respectively, and do the same for $\mathcal{Q}_{2}$.
Eq.\eqref{eq:ncm} minimizes the differences between two views of instances of the same category.
The whole process is illustrated in \figurename~\ref{fig:overview}.
Therefore, we computed the cross-view classification loss as follows:
\begin{equation}
\footnotesize
    L_{meta} = \frac{1}{4} \sum_{m,n} L_{mn}.
\label{Lmeta}
\end{equation}
\noindent\textbf{Distance-scaled Contrastive Loss.}
Since contrastive learning approaches work solely at the instance level, it is superficial to simply add contrastive loss into meta-training without taking full
advantage of the episodic training mechanism catered for FSL.
To better apply contrastive learning to the meta-training, we perform query instance discrimination between two views of the shared episodes.
Specifically, inspired by \cite{SimCLRv2}, we further inherit the pre-trained projection head $proj(\cdot)$ in Sect.~\ref{sec:pretrain} to map each sample $x_i$ in the two episodes $E_1$ and $E_2$ into a projected vector $\mathbf{z}_{i} \in \mathbb{R}^{D}$.
We similarly obtain the prototype $\mathbf{o}_{r}^{k}$ by averaging the projected vector of the same classes, that is
$\mathbf{o}_{r}^{k} = \frac{1}{|\mathcal{S}_{r}^{k}|} \sum_{(\mathbf{z}_i,y_i)\in \mathcal{S}_{r}^{k}} \mathbf{z}_i$ where $r$ is the same as in Eq.~\eqref{eq:meta_prototype}.
For each query vector $\mathbf{z}_{i}$ in $\mathcal{Q}_1$ and $\mathcal{Q}_2$,
we reconstruct its positive sample set by using corresponding augmented version $\mathbf{z}_{i}^{\prime}$ and the samples of class $y_{i}$ in both support sets.
Therefore, we reformulate the supervised contrastive loss \cite{SupContrast} in the form of episodic training as follows:
\begin{equation}
\footnotesize
    {L}(\mathbf{z}_i)=
    - \sum_{\mathbf{z}_{H} \in H(\mathbf{z}_i)} \log \frac{\lambda_{\mathbf{z}_{i}\mathbf{z}_{H}} \exp \left(\mathbf{z}_{i} \cdot \mathbf{z}_{H} / \tau_{5}\right)}{\sum_{\mathbf{z}_{A} \in A(\mathbf{z}_i)} \lambda_{\mathbf{z}_i \mathbf{z}_{A}} \exp \left(\mathbf{z}_{i} \cdot \mathbf{z}_{A} / \tau_{5}\right)}.
\end{equation}
Here, operation $\cdot$ means the inner product between features after $l_{2}$ normalization.
$\tau_{5}$ is a temperature parameter.
The $\lambda_{\mathbf{z}_{i}\mathbf{z}_{j}}=2-dist(\mathbf{z}_{i}, \mathbf{z}_{j})$ is the coefficient that reflects the distance relationship between $\mathbf{z}_{i}$ and $\mathbf{z}_{j}$, where $dist(\cdot,\cdot)$ refers to cosine similarity.
$H(\mathbf{z}_i)=\left\{\mathbf{z}_{i}^{\prime}\right\} \cup \mathcal{S}_{1}^{y_{i}} \cup \mathcal{S}_{2}^{y_{i}}$ is the positive set of $\mathbf{z}_{i}$,
and $A(\mathbf{z}_i)=\left\{\mathbf{z}_{i}^{\prime}\right\} \cup \mathcal{S}_{1} \cup \mathcal{S}_{2} \cup \left\{\mathbf{o}_{1}^{k}\right\}^{M}_{k=1}\cup \left\{\mathbf{o}_{2}^{k}\right\}^{M}_{k=1}$.
The distance coefficient $\lambda_{\mathbf{z}_{i}\mathbf{z}_{j}}$ and additional prototypes $\mathbf{o}_{r}^{k}$ on both views of the original episode are introduced to reduce the similarities of queries to their positives.
Thus the model can learn more discriminative representations adapted to different tasks.
Then we compute the distance-scaled contrastive loss as:
\begin{equation}
\footnotesize
    {L}_{info}=\sum_{\mathbf{z}_i \in \mathcal{Q}_{1} \cup  \mathcal{Q}_{2}} \frac{1}{|H(\mathbf{z}_i)|} {L}(\mathbf{z}_i).
\label{Linfo}
\end{equation}
By optimizing $L_{info}$, we force the representations to capture information about instance discrimination episodically and learn the interrelationships among samples of the same category in cross-view episodes.

\noindent\textbf{Objective in Meta-training.}
In meta-training, we mainly build two losses as in Eq.~\eqref{Lmeta} and Eq.~\eqref{Linfo} in order to enhance the transferability and discriminative ability of representations.
Then the total objective function in meta-training is defined as:
\begin{equation}
\footnotesize
    L_{total} = L_{meta} + \beta L_{info},
\end{equation}
where $\beta$ is a balance scalar and will be detailed in the following section.

\section{Experiments}
\label{sec:experiment}

\subsection{Datasets and Setup}
\textbf{Datasets.}
We evaluate our method on three popular benchmark datasets.
The \emph{miniImageNet} dataset \cite{MatchNet} contains 100 classes with 600 images per class, and these classes are divided into 64, 16 and 20 for the training, validation and test sets, respectively.
The \emph{tieredImageNet} dataset \cite{tiered} contains 608 classes grouped in 34 high-level categories with 779,165 images, where 351 classes are used for training, 97 for validation and 160 for testing.
The \emph{CIFAR-FS} dataset \cite{cifar-fs} contains 100 classes with 600 images per class.
These classes are split into the training, validation and test sets in proportions of 64, 16 and 20.

\noindent\textbf{Implementation Details.}
Following the baseline \cite{FEAT}, we use ResNet-12 as backbone and a multihead attention as the task-specific module (number of heads is 1).
All projection heads in our method has the same structure as in \cite{SimCLR}.
We adopt SGD optimizer with a weight decay of 5e-4 and a momentum of 0.9 for both the pre-training and meta-training stages.
During pre-training, the learning rate is initialized to be 0.1 and adapted via cosine learning rate scheduler after warming up.
Temperature parameters $\tau_{1,2,3,4}$ are set to 0.1 and the balance scalars $\alpha_{1,2,3}$ are set to 1.0.
During meta-training, temperature $\tau_{5}$ are set to 0.1.
We use StepLR with a step size of 40 and gamma of 0.5 and set $\beta=0.01$ for 1-shot.
For 5-shot, StepLR is used with a step size of 50 and gamma of 0.5, and the $\beta$ is set to 0.1.

\noindent\textbf{Evaluation.}
We follow the 5-way 1-shot and 5-way 5-shot few-shot classification tasks.
In meta-testing, our method (inductive) simply classify the query samples by computing the euclidean distance between the prototypes and query samples.
We randomly sample 2000 episodes from test set in meta-testing with 15 query images per class and report the mean accuracy together with corresponding 95\% confidence interval.

\noindent\textbf{Data Augmentation.} For all datasets,
we empirically find that standard \cite{FEAT,DeepEMD} and SimCLR-style \cite{SimCLR,MoCo} data augmentation strategies work best in pre-training and meta-training.
During meta-testing, no data augmentation strategy is used.
The image transformations used in standard strategy include randomresizedcrop, colorJitter and randomhorizontalflip, while SimCLR-style strategy contains randomresizedcrop, randomhorizontalflip, randomcolorjitter and randomgrayscale.

\begin{table}[!t]
    \centering
\caption{The average 5-way few-shot classification accuracies(\%) with 95\% confidence interval on miniImageNet and tieredImageNet.}
    \renewcommand\arraystretch{1.1}
\resizebox{0.95\columnwidth}{!}{
\begin{tabular}{lccccc}
\hline
\multicolumn{1}{l|}{}                                   & \multicolumn{1}{c|}{}                                    & \multicolumn{2}{c}{\textbf{miniImageNet}}     & \multicolumn{2}{c}{\textbf{tieredImageNet}}       \\ \cline{3-6} 
\multicolumn{1}{l|}{\multirow{-2}{*}{\textbf{Method}}} & \multicolumn{1}{c|}{\multirow{-2}{*}{\textbf{Backbone}}} & \textbf{1-shot}       & \textbf{5-shot}       & \textbf{1-shot}         & \textbf{5-shot}         \\ \hline
\multicolumn{1}{l|}{MAML~\cite{MAML}}                               & \multicolumn{1}{c|}{}                                    & 48.70 ± 1.75          & 63.11 ± 0.92          & ---                       & ---                       \\
\multicolumn{1}{l|}{RelationNets~\cite{RelationNet}}                       & \multicolumn{1}{c|}{}                                    & 50.44 ± 0.82          & 65.32 ± 0.70          & 54.48 ± 0.93            & 71.32 ± 0.78            \\
\multicolumn{1}{l|}{MatchingNets~\cite{MatchNet}}                       & \multicolumn{1}{c|}{}                                    & 48.14 ± 0.78          & 63.48 ± 0.66          & ---                       & ---                       \\
\multicolumn{1}{l|}{ProtoNets~\cite{ProtoNet}}                          & \multicolumn{1}{c|}{\multirow{-4}{*}{CONV-4}}             & 44.42 ± 0.84          & 64.24 ± 0.72          & 53.31 ± 0.89            & 72.69 ± 0.74            \\ \hline
\multicolumn{1}{l|}{LEO~\cite{LEO}}                                & \multicolumn{1}{c|}{}                                    & 61.76 ± 0.08          & 77.59 ± 0.12          & 66.33 ± 0.05            & 82.06 ± 0.08            \\
\multicolumn{1}{l|}{CC+rot~\cite{Boost}}      & \multicolumn{1}{c|}{}                                    & 62.93 ± 0.45          & 79.87 ± 0.33          & 62.93 ± 0.45            & 79.87 ± 0.33            \\
\multicolumn{1}{l|}{wDAE~\cite{wDAE}}        & \multicolumn{1}{c|}{}         & 61.07 ± 0.15          & 76.75 ± 0.11          & 68.18 ± 0.16            & 83.09 ± 0.12            \\
\multicolumn{1}{l|}{PSST~\cite{PSST}}                                & \multicolumn{1}{c|}{\multirow{-3}{*}{WRN-28-10}}                                    & 64.16 ± 0.44          & 80.64 ± 0.32          & ---            & ---            \\\hline
\multicolumn{1}{l|}{TADAM~\cite{TADAM}}                              & \multicolumn{1}{c|}{}                                    & 58.5 ± 0.3            & 76.7 ± 0.3            & ---                       & ---                       \\
\multicolumn{1}{l|}{MetaOptNet~\cite{MetaOptNet}}                         & \multicolumn{1}{c|}{}                                    & 62.64 ± 0.61          & 78.63 ± 0.46          & 65.99 ± 0.72            & 81.56 ± 0.53            \\
\multicolumn{1}{l|}{DeepEMD~\cite{DeepEMD}}                            & \multicolumn{1}{c|}{}                                    & 65.91 ± 0.82          & 82.41 ± 0.56          & 71.16 ± 0.87            & 86.03 ± 0.58            \\
\multicolumn{1}{l|}{CAN~\cite{CAN}}                                & \multicolumn{1}{c|}{}                                    & 63.85 ± 0.48          & 79.44 ± 0.34          & 69.89 ± 0.51            & 84.23 ± 0.37            \\
\multicolumn{1}{l|}{FEAT~\cite{FEAT}}                               & \multicolumn{1}{c|}{}                                    & 66.78 ± 0.20          & 82.05 ± 0.14          & 70.80 ± 0.23            & 84.79 ± 0.16            \\
\multicolumn{1}{l|}{RFS~\cite{rfs}}                     & \multicolumn{1}{c|}{}                                    & 62.02 ± 0.63          & 79.64 ± 0.44          & 69.74 ± 0.72            & 84.41 ± 0.55            \\
\multicolumn{1}{l|}{InfoPatch~\cite{InfoPatch}}                          & \multicolumn{1}{c|}{}                                    & 67.67 ± 0.45          & 82.44 ± 0.31          & 71.51 ± 0.52            & 85.44 ± 0.35            \\
\multicolumn{1}{l|}{DMF~\cite{DMF}}                                & \multicolumn{1}{c|}{}                                    & 67.76 ± 0.46          & 82.71 ± 0.31          & 71.89 ± 0.52            & 85.96 ± 0.35            \\
\multicolumn{1}{l|}{RENet~\cite{RENet}}                              & \multicolumn{1}{c|}{}                                    & 67.60 ± 0.44          & 82.58 ± 0.30          & 71.61 ± 0.51            & 85.28 ± 0.35            \\
\multicolumn{1}{l|}{BML~\cite{BML}}                                & \multicolumn{1}{c|}{}                                    & 67.04 ± 0.63          & 83.63 ± 0.29          & 68.99 ± 0.50            & 85.49 ± 0.34            \\
\multicolumn{1}{l|}{PAL~\cite{PAL}}                                & \multicolumn{1}{c|}{}                                    & 69.37 ± 0.64          & 84.40 ± 0.44          & 72.25 ± 0.72            & \textbf{86.95 ± 0.47}            \\
\multicolumn{1}{l|}{TPMN~\cite{TPMN}}                               & \multicolumn{1}{c|}{\multirow{-12}{*}{ResNet-12}}        & 67.64 ± 0.63          & 83.44 ± 0.43          & 72.24 ± 0.70            & 86.55 ± 0.63            \\ \hline
\multicolumn{1}{l|}{Ours}                               & \multicolumn{1}{c|}{ResNet-12}                           & \textbf{70.19 ± 0.46} & \textbf{84.66 ± 0.29} & \textbf{72.62 ± 0.51} & 86.62 ± 0.33 \\ \hline
\end{tabular}
}
\label{table:mini_tiered}
\end{table}

\subsection{Main Results}
In this subsection, we compare our method with competitors on three mainstream FSL datasets and the results are reported in \tablename~\ref{table:mini_tiered} and \tablename~\ref{table:cifar_fs}.
We can observe that our proposed method consistently achieves competitive results compared to the current state-of-the-art (SOTA) FSL methods on both the 5-way 1-shot and 5-way 5-shot tasks.
For the miniImageNet dataset (\tablename~\ref{table:mini_tiered}), our proposed method outperforms the current best results by 0.82\% in the 1-shot task and 0.26\% in 5-shot task.
For the tieredImageNet dataset (\tablename~\ref{table:mini_tiered}), our method improves over the current SOTA method by 0.37\% for 1-shot and achieves the second best 5-shot result.
Note that our method outperforms the original FEAT by 1.82\% for 1-shot and 1.83\% for 5-shot on tieredImageNet.
For the CIFAR-FS dataset (\tablename~\ref{table:cifar_fs}),
our method surpasses the current SOTA by 0.46\% and 0.19\% in the 1-shot and 5-shot tasks, respectively.
Compared to those methods \cite{PSST,Boost,InfoPatch,PAL}, our proposed method works better on most datasets.
The consistent and competitive results on the three datasets indicate that our method can learn more transferable representations by incorporating contrastive learning in both the pre-training and meta-training stages.

\begin{table}[t]
\centering
\caption{The average 5-way few-shot classification accuracies(\%) with 95\% confidence interval on CIFAR-FS. $^{\star}$ results used our implementation.}
\renewcommand\arraystretch{1.1}
\resizebox{0.72\columnwidth}{!}{
\begin{tabular}{l|c|cc}
\hline
\textbf{Method}                   & \textbf{Backbone}          & \textbf{1-shot} & \textbf{5-shot} \\ \hline
Ravichandran \emph{et al.}~\cite{Ravichandran} &               \multirow{2}{*}{CONV-4}             & 55.14 ± 0.48    & 71.66 ± 0.39    \\
ConstellationNet~\cite{constellation}                  &     & 69.3 ± 0.3      & 82.7 ± 0.2      \\ \hline
CC+rot~\cite{Boost}                            & \multirow{2}{*}{WRN-28-10} & 75.38 ± 0.31    & 87.25 ± 0.21    \\
PSST~\cite{PSST}                              &                            & 77.02 ± 0.38    & 88.45 ± 0.35    \\ \hline
Ravichandran \emph{et al.}~\cite{Ravichandran} & \multirow{9}{*}{ResNet-12} & 69.15 ± -       & 84.7 ± -        \\
MetaOptNet~\cite{MetaOptNet}                        &                            & 72.0 ± 0.7      & 84.2 ± 0.5      \\
Kim \emph{et al.}~\cite{KimKK20a}          &                            & 73.51 ± 0.92    & 85.65 ± 0.65    \\
FEAT$^{\star}$~\cite{FEAT}                               &                            & 75.41 ± 0.21      & 87.32 ± 0.15      \\
RFS~\cite{rfs}                               &                            & 73.9 ± 0.8      & 86.9 ± 0.5      \\
ConstellationNet~\cite{constellation}                  &                            & 75.4 ± 0.2      & 86.8 ± 0.2      \\
RENet~\cite{RENet}                             &                            & 74.51 ± 0.46    & 86.60 ± 0.32    \\
BML~\cite{BML}                               &                            & 73.45 ± 0.47    & 88.04 ± 0.33    \\
PAL~\cite{PAL}                               &                            & 77.1 ± 0.7      & 88.0 ± 0.5      \\
TPMN~\cite{TPMN}                              &                            & 75.5 ± 0.9      & 87.2 ± 0.6      \\ \hline
Ours  &  ResNet-12    & \textbf{77.56 ± 0.47} & \textbf{88.64 ± 0.31} \\
\hline
\end{tabular}
}
\label{table:cifar_fs}
\end{table}

\subsection{Ablation Study}
In this subsection, we study the effectiveness of different components in our method on three datasets.
The results in \tablename~\ref{table:ablation} show a significant improvement in the performance of our proposed method compared to the baseline pre-trained by CE loss ($L_{CE}$) only.
Specifically, using $L_{CE}$ and all contrastive losses $L_{CL}$ in pre-training improves the accuracy by an average of 1.66\% (1-shot) and 1.29\% (5-shot) on the three datasets.
This allows the representations to generalize better, rather than just focusing on the information only needed for the classification on base classes.
We then apply the CVET and $L_{info}$ respectively, and the consistent improvements suggest that CVET and $L_{info}$ are effective.
The results jointly based on CVET and $L_{info}$ are further enhanced by 0.95\% (1-shot) and 0.72\% (5-shot) in average, indicating that our full method increases the transferability of representations on novel classes.
Note that the gains obtained by CVET and $L_{info}$ are relatively low on miniImageNet, as the number of training steps on this dataset is very small,
leading to a sufficiently narrow gap between two different views of the same image at the end of pre-training.
Additionally, more ablation experiments on the effectiveness of all parts in $L_{CL}$ are included in the supplementary material.

\begin{table}[!t]
\centering
\caption{Ablation experiments on miniImageNet. $L_{CE}$ means the baseline model is pre-trained with CE loss. $L_{CL}$ denotes all contrastive losses used in pre-training. CVET and $L_{info}$ are only available in meta-training.}
\renewcommand\arraystretch{1.1}
\resizebox{0.99\columnwidth}{!}{
\begin{tabular}{cccc|ll|ll|ll}
\hline
\multirow{2}{*}{$L_{CE}$} & \multirow{2}{*}{$L_{CL}$} & \multirow{2}{*}{CVET} & \multirow{2}{*}{$L_{info}$} & \multicolumn{2}{c|}{\textbf{miniImageNet}}                                  & \multicolumn{2}{c|}{\textbf{tieredImageNet}}                                & \multicolumn{2}{c}{\textbf{CIFAR-FS}}                                      \\ \cline{5-10} 
                          &                           &                       &                             & \multicolumn{1}{c|}{\textbf{1-shot}} & \multicolumn{1}{c|}{\textbf{5-shot}} & \multicolumn{1}{c|}{\textbf{1-shot}} & \multicolumn{1}{c|}{\textbf{5-shot}} & \multicolumn{1}{c|}{\textbf{1-shot}} & \multicolumn{1}{c}{\textbf{5-shot}} \\ \hline
$\surd$                   &                           &                       &                             & \multicolumn{1}{l|}{66.58 ± 0.46}    & 81.92 ± 0.31                         & \multicolumn{1}{l|}{70.41 ± 0.51}    & 84.69 ± 0.36                         & \multicolumn{1}{l|}{75.54 ± 0.48}    & 87.28 ± 0.32                        \\
$\surd$                   & $\surd$                   &                       &                             & \multicolumn{1}{l|}{69.53 ± 0.47}    & 84.33 ± 0.29                         & \multicolumn{1}{l|}{71.83 ± 0.51}    & 85.64 ± 0.35                         & \multicolumn{1}{l|}{76.15 ± 0.47}    & 87.79 ± 0.33                        \\
$\surd$                   & $\surd$                   & $\surd$               &                             & \multicolumn{1}{l|}{69.89 ± 0.46}    & 84.43 ± 0.29                         & \multicolumn{1}{l|}{72.39 ± 0.52}    & 86.07 ± 0.35                         & \multicolumn{1}{l|}{77.02 ± 0.48}    & 88.28 ± 0.32                        \\
$\surd$                   & $\surd$                   &                       & $\surd$                     & \multicolumn{1}{l|}{69.78 ± 0.46}    & 84.46 ± 0.29                         & \multicolumn{1}{l|}{72.51 ± 0.52}    & 86.23 ± 0.34                         & \multicolumn{1}{l|}{77.37 ± 0.49}    & 88.48 ± 0.32                        \\
$\surd$                   & $\surd$                   & $\surd$               & $\surd$                     & \multicolumn{1}{l|}{70.19 ± 0.46}    & 84.66 ± 0.29                         & \multicolumn{1}{l|}{72.62 ± 0.51}    & 86.62 ± 0.33                         & \multicolumn{1}{l|}{77.56 ± 0.47}    & 88.64 ± 0.31                        \\ \hline
\end{tabular}
}
\label{table:ablation}
\end{table}

\begin{table}[!t]
\centering
\caption{Comparison experiments of different data augmentation strategies on miniImageNet.}
\renewcommand\arraystretch{1.1}
\resizebox{0.58\columnwidth}{!}{
\begin{tabular}{l|c|c}
\hline
\textbf{Augmentation}     & \textbf{1-shot}       & \textbf{5-shot}       \\ \hline
Standard~\cite{FEAT,DeepEMD}         & 67.84 ± 0.47 & 82.82 ± 0.31 \\
SimCLR-style~\cite{SimCLR,MoCo}           & 70.19 ± 0.46 & 84.66 ± 0.29 \\
AutoAugment~\cite{AutoAugment}      & 68.45 ± 0.46 & 83.94 ± 0.30 \\
StackedRandAug~\cite{StackedRandAug}  & 68.31 ± 0.46 & 83.13 ± 0.31 \\ \hline
\end{tabular}
}
\label{table:augmentation}
\end{table}

\subsection{Further Analysis}
\label{sec:further}
\noindent
\textbf{Data augmentation.}
We investigate the impact of employing a different data augmentation strategy while maintaining the use of the standard one.
\tablename~\ref{table:augmentation} shows that the SimCLR-style \cite{SimCLR,MoCo} strategy works best while the AutoAugment \cite{AutoAugment} and StackedRandAug\cite{StackedRandAug} perform slightly worse.
However, all three strategies are better than the standard one.
We believe that the standard strategy lacks effective random image transformations and thus always produces two similar views of the same image during the training process.
Conversely, the AutoAugment and StackedRandAug strategies that intensely change the original image may lead to excessive bias between two views of the same image, thereby increasing the difficulty of performing contrastive learning.

\begin{figure}[!t]
    \centering
    \begin{minipage}[t]{0.24\columnwidth}
		\centering
        \includegraphics[width=1\linewidth]{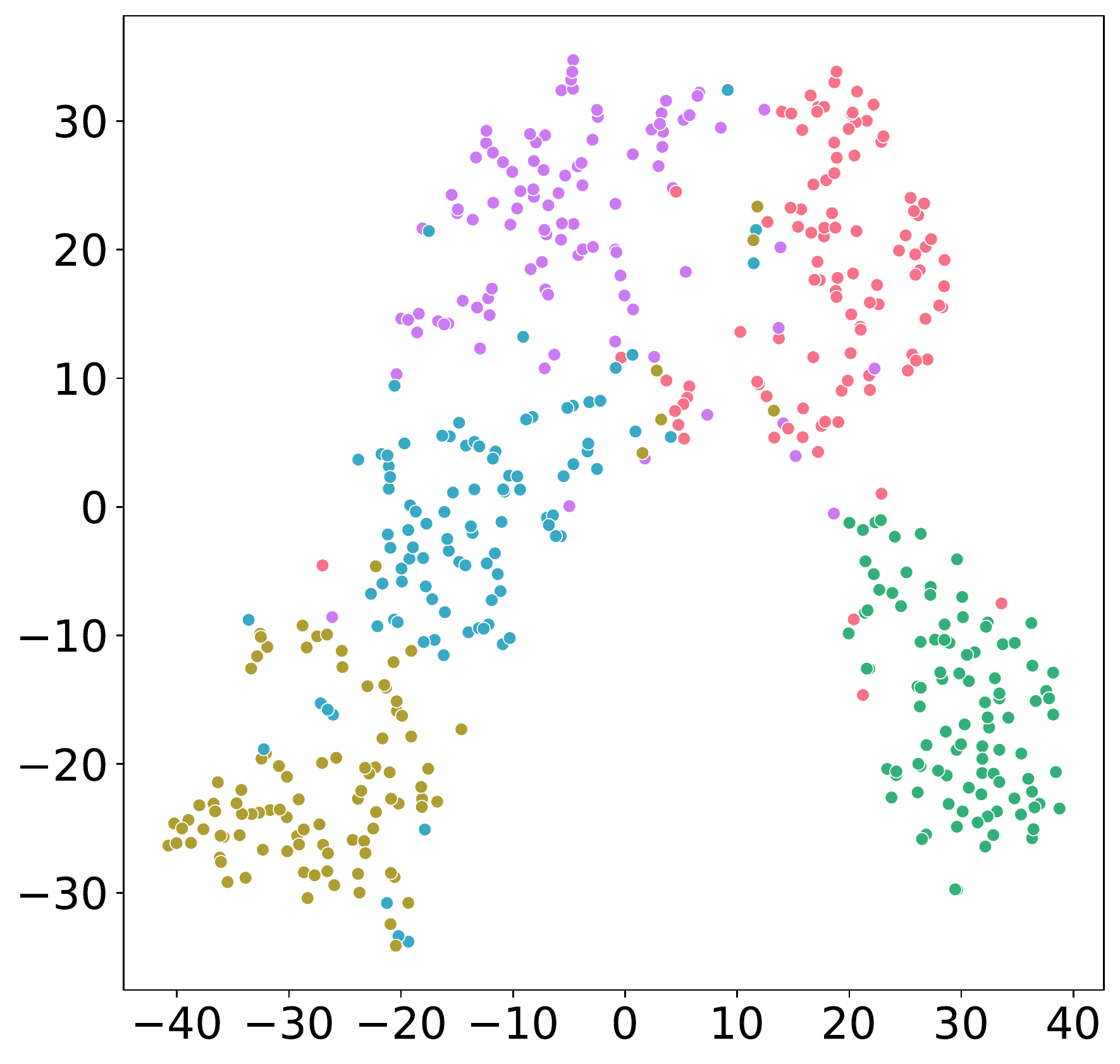}
        \subcaption{ProtoNet}
		\label{fig:visual_protonet}
	\end{minipage}
	\begin{minipage}[t]{0.24\columnwidth}
		\centering
        \includegraphics[width=1\linewidth]{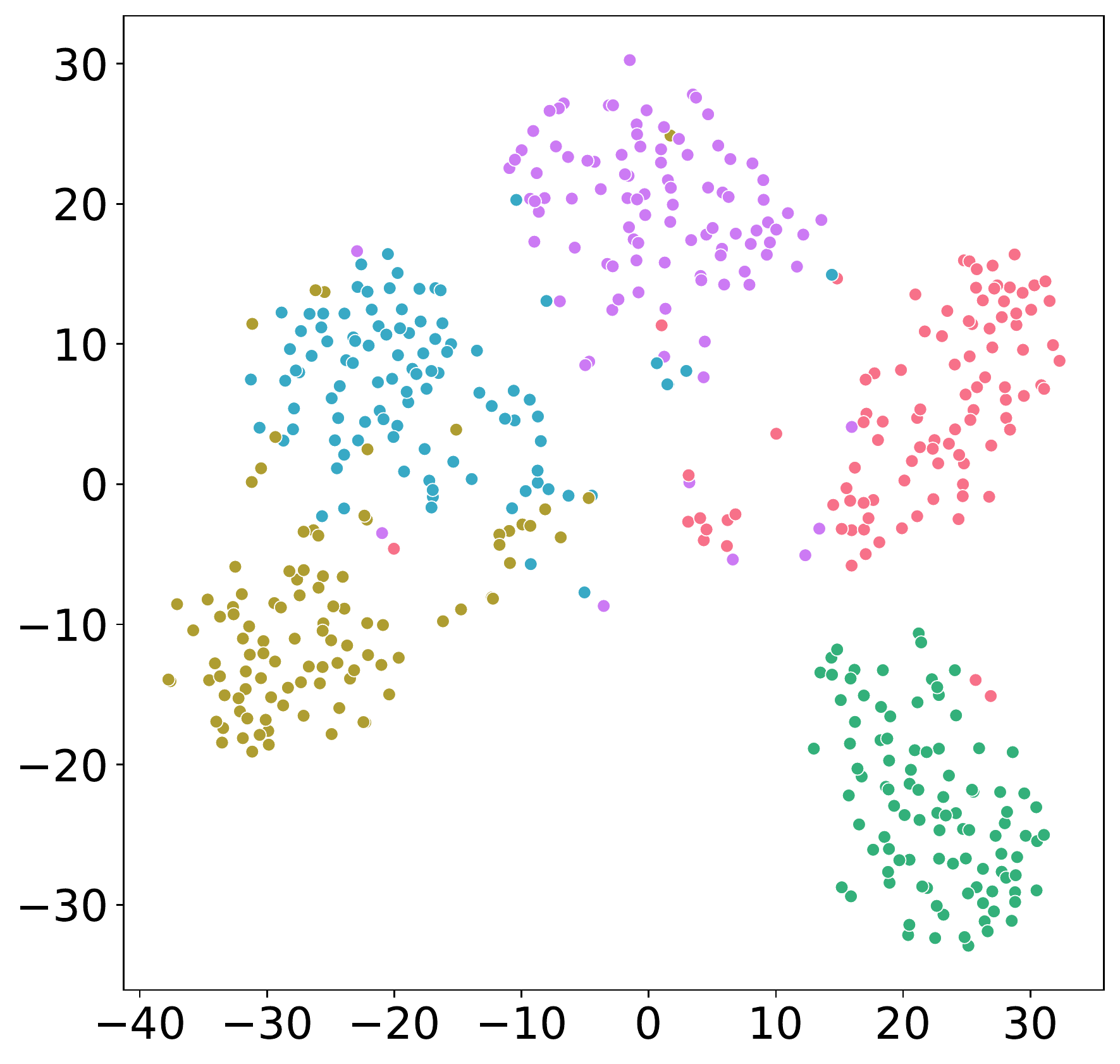}
        \subcaption{ProtoNet+Ours}
		\label{fig:visual_protonet_ours}
	\end{minipage}
    \begin{minipage}[t]{0.24\columnwidth}
		\centering
        \includegraphics[width=1\linewidth]{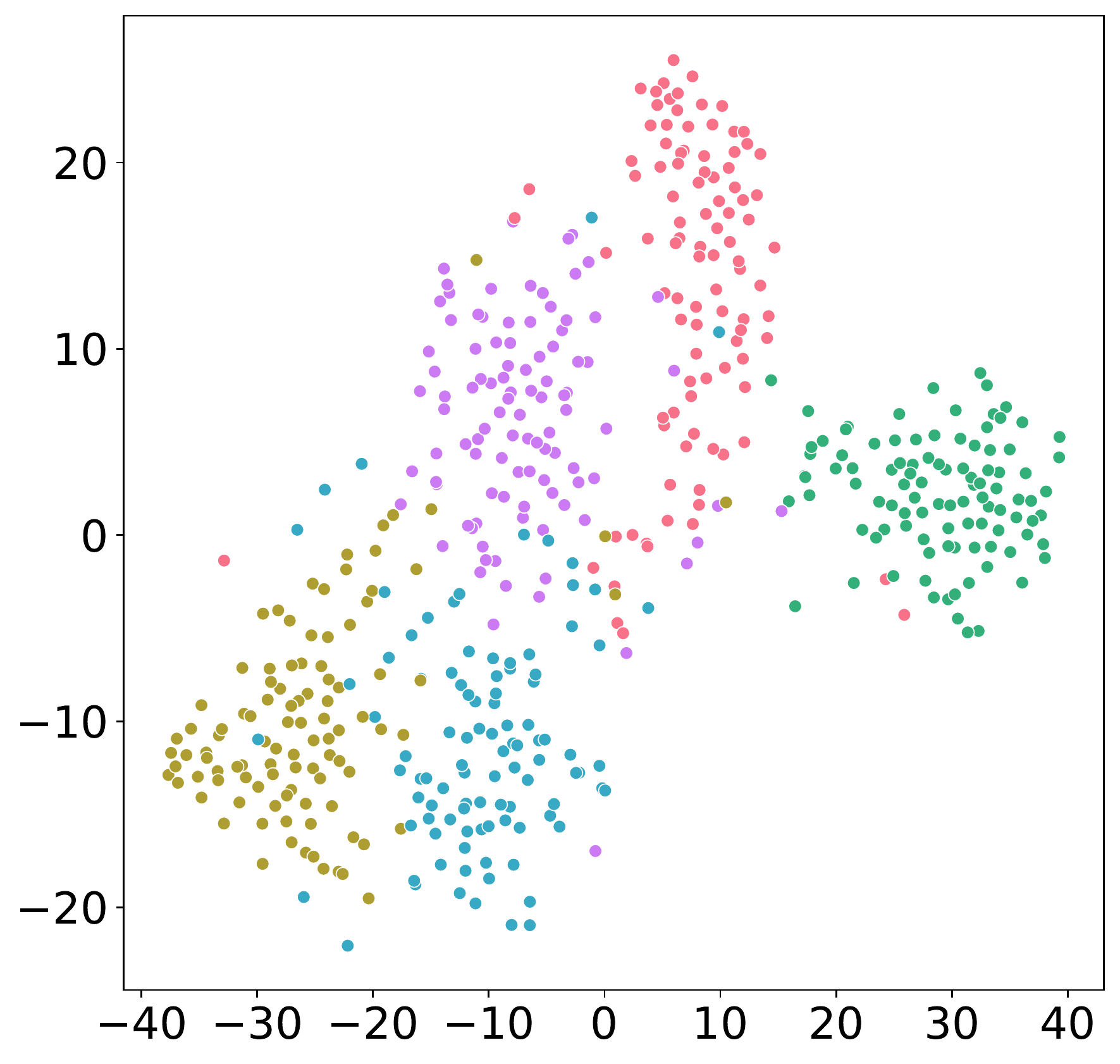}
        \subcaption{baseline}
		\label{fig:visual_baseline}
	\end{minipage}
    \begin{minipage}[t]{0.24\columnwidth}
		\centering
        \includegraphics[width=1\linewidth]{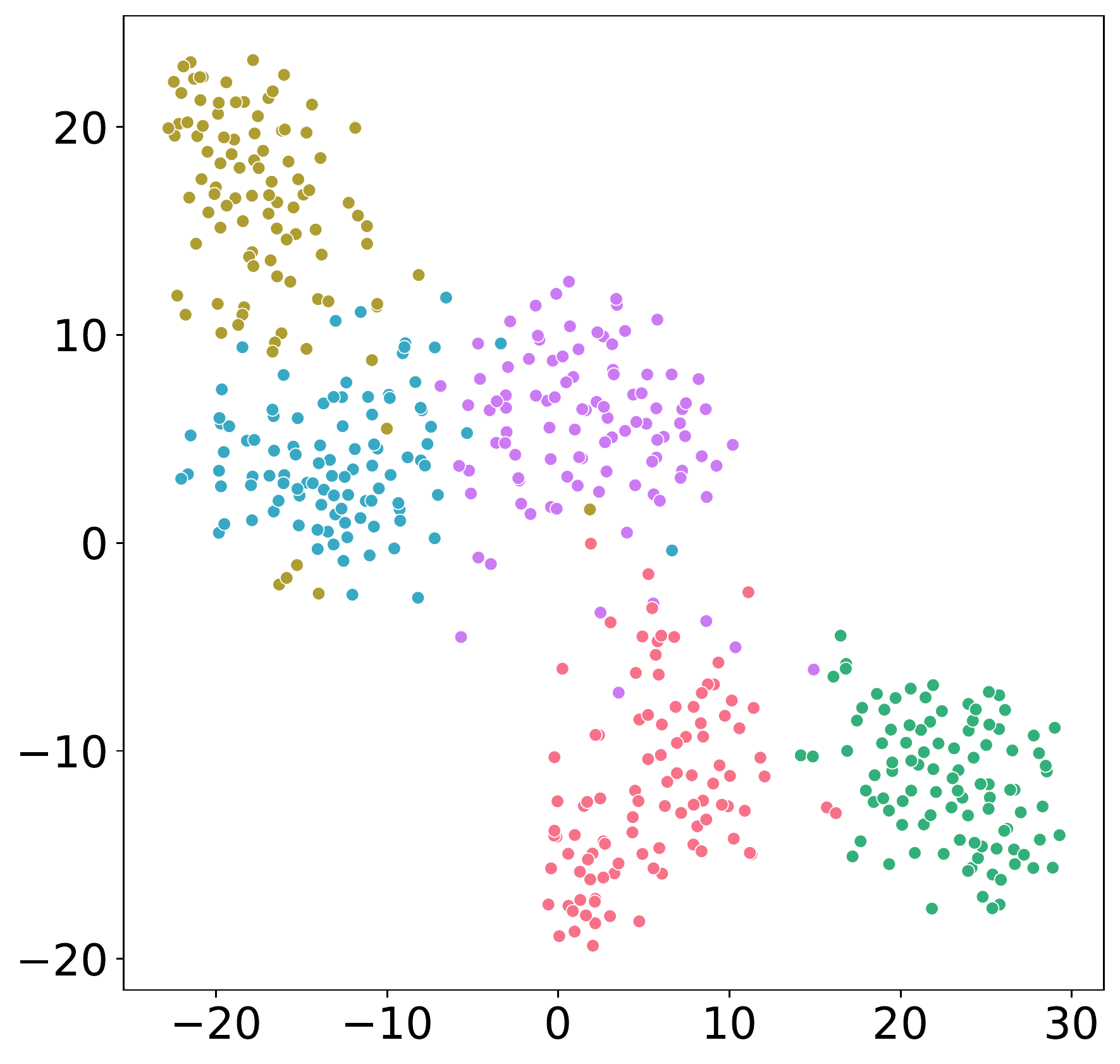}
        \subcaption{Ours}
		\label{fig:visual_ours}
	\end{minipage}
\caption{Visualization of 100 randomly sampled images for each of the 5 meta-test classes from miniImageNet by t-SNE~\cite{tsne}.}
\label{fig:visual_experiment}
\end{figure}

\noindent\textbf{Visualization.}
We give visualization to validate the transferability of representations produced by our framework on novel classes.
Specifically, we randomly sample 5 classes from the meta-test set of miniImageNet with 100 images per class and obtain embeddings of all images using ProtoNet~\cite{ProtoNet}, ProtoNet+Ours, baseline and Ours (baseline + our framework), respectively.
Then we use t-SNE \cite{tsne} to project these embeddings into 2-dimensional space as shown in \figurename~\ref{fig:visual_experiment}.
The distributions of the embeddings obtained by ProtoNet + Ours and baseline + our framework are more separable and the class boundaries more precise and compact (see \figurename~\ref{fig:visual_protonet_ours}, \figurename~\ref{fig:visual_ours} $vs.$ \figurename~\ref{fig:visual_protonet}, \figurename~\ref{fig:visual_baseline}).
The visualization results indicate that our proposed framework generates more transferable and discriminative representations on novel classes.
The complete visualization results are available in the supplementary material.

\begin{table}[!t]
\centering
\caption{Method combinations on miniImageNet.}
\renewcommand\arraystretch{1.1}
\resizebox{0.48\columnwidth}{!}{
\begin{tabular}{lc|c}
\hline
\multicolumn{1}{l|}{\textbf{Model}}         & \textbf{1-shot}       & \textbf{5-shot}       \\ \hline
\multicolumn{1}{l|}{ProtoNet}      & 63.77 ± 0.47 & 80.58 ± 0.32 \\
\multicolumn{1}{l|}{ProtoNet+CL}   & 66.17 ± 0.46 & 81.73 ± 0.30 \\
\multicolumn{1}{l|}{ProtoNet+Ours} & 66.51 ± 0.47 & 81.97 ± 0.30 \\
\multicolumn{1}{l|}{baseline}      & 66.23 ± 0.47 & 80.81 ± 0.33 \\
\multicolumn{1}{l|}{baseline+CL}   & 69.53 ± 0.47 & 84.33 ± 0.29 \\
\multicolumn{1}{l|}{Ours} & 70.19 ± 0.46 & 84.66 ± 0.29 \\ \hline
\end{tabular}
}
\label{table:transferability}
\end{table}
\noindent\textbf{Method Combination.}
We combine our proposed framework with the two-stage FSL methods ProtoNet~\cite{ProtoNet} and baseline (FEAT~\cite{FEAT}), respectively.
The results are shown in \tablename~\ref{table:transferability}.
Integration with our framework improves the accuracy of ProtoNet by 2.74\% (1-shot) and 1.39\% (5-shot).
Similarly, our framework improves the 1-shot accuracy of baseline by 3.96\% and its 5-shot accuracy by 3.85\%.
The results in \figurename~\ref{fig:visual_experiment} and \tablename~\ref{table:transferability} show that our framework can be applied to other two-stage FSL methods and effectively enhance the performance.

\section{Conclusions}
In this paper,
we apply contrastive learning to the two-stage training paradigm of FSL to alleviate the limitation on generalizability of representations.
Concretely, we use vector-map and map-map modules to incorporate self-supervised contrastive losses in the pre-training stage.
We further propose a CVET strategy and a distance-scaled contrastive loss to extend contrastive learning to the meta-training stage effectively.
The comprehensive experimental results show that our proposed method achieves competitive performance on three well-known FSL datasets miniImageNet, tieredImageNet and CIFAR-FS.
\\
~\\
\noindent\textbf{Acknowledgement.}
This work are supported by:
(i) National Natural Science Foundation of China (Grant No. 62072318 and No. 62172285);
(ii) Natural Science Foundation of Guangdong Province of China (Grant No. 2021A1515012014);
(iii) Science and Technology Planning Project of Shenzhen Municipality (Grant No. JCYJ20190808172007500).

%
%
\bibliographystyle{splncs04}
\bibliography{egbib}

\clearpage

\appendix
\centerline{\LARGE\bf Supplementary Material}
\vspace{1mm}

\section{Additional Ablation Study}
We conduct ablation studies in the pre-training stage on the miniImageNet.
Here, except for the model trained with only $L_{CE}$ (see the first line in \tablename~\ref{tab:pretrain_loss}), we adopt cross-view episodic training (CVET) mechanism and distance-scaled contrastive loss in the meta-training stage for all experiments in \tablename~\ref{tab:pretrain_loss} and \tablename~\ref{tab:pretrain_local}.

\vspace{-15pt}

\begin{table*}[!ht]
\centering
\caption{Ablation experiments in the pre-training stage on miniImageNet.}
\renewcommand\arraystretch{1.2}
\begin{tabular}{c|ccc|c|c}
\hline
$L_{CE}$                & $L_{global}^{ss}$ & $L_{local}^{ss}$ & $L_{global}^{s}$ & \textbf{1-shot}       & \textbf{5-shot}        \\ \hline
$\surd$ &                                       &                                      &                                      & 66.58 ± 0.46          & 81.92 ± 0.31          \\
$\surd$ & $\surd$                  &                                      &                                      & 68.53 ± 0.46 & 83.86 ± 0.29 \\
$\surd$ &                                       & $\surd$                 &                                      & 69.33 ± 0.46 & 84.02 ± 0.30 \\
$\surd$ &                                       &                                      & $\surd$                 & 67.88 ± 0.45 & 83.37 ± 0.29 \\
$\surd$ & $\surd$                  & $\surd$                 &                                      & 69.03 ± 0.46 & 83.87 ± 0.30 \\
$\surd$ & $\surd$                  &                                      & $\surd$                 & 68.68 ± 0.46 & 84.17 ± 0.29 \\
$\surd$ &                                       & $\surd$                 & $\surd$                 & 69.72 ± 0.45 & 84.49 ± 0.29 \\
$\surd$ & $\surd$                  & $\surd$                 & $\surd$                 & 70.19 ± 0.46 & 84.66 ± 0.29 \\ \hline
\end{tabular}
\label{tab:pretrain_loss}
\end{table*}

\vspace{-25pt}

\begin{table*}[!ht]
\centering
\caption{Effectiveness of vector-map and map-map modules on miniImageNet.}
\renewcommand\arraystretch{1.5}
\begin{tabular}{c|cc|c|c}
\hline
$L_{CE}+L_{global}^{ss}+L_{global}^{s}$ & $L_{vec-map}^{ss}$ & $L_{map-map}^{ss}$ & \textbf{1-shot} & \textbf{5-shot} \\\hline
$\surd$     &            &                &  68.68 ± 0.46   &  84.17 ± 0.29   \\
$\surd$     & $\surd$           &                & 69.29 ± 0.46    & 84.23 ± 0.29    \\
$\surd$     &                & $\surd$           & 68.81 ± 0.46    & 84.25 ± 0.29    \\
$\surd$     &  $\surd$              & $\surd$           & 70.19 ± 0.46    & 84.66 ± 0.29    \\ \hline
\end{tabular}
\label{tab:pretrain_local}
\end{table*}

As shown in \tablename~\ref{tab:pretrain_loss}, compared to training with $L_{CE}$ alone, each contrastive loss used in the pre-training stage plays an important role, with $L_{local}^{ss}$ contributing the most.
The results from the methods introducing $L_{local}^{ss}$ indicate that contrastive learning based on extra local information can learn more generalizable representations.
Meanwhile, we can observe that the results obtained by training with supervision only are much lower than the results obtained by using both supervision and self-supervision.
The results in \tablename~\ref{tab:pretrain_loss} validate the effectiveness of our proposed contrastive losses, and we obtain the best results when employing all proposed contrastive losses in the pre-training stage.

Based on $L_{CE}$ and the other two contrastive losses using global information, we verify the effectiveness of vector-map and map-map modules by conducting experiments on them separately, as shown in \tablename~\ref{tab:pretrain_local}.
Compared to using only global information, contrastive learning that leverages local information either in the form of vector-map or map-map can improve the transferability of the representations.
The best results are achieved when both forms work together.

\newpage

\section{Experiments on Hyperparameters in the Pre-training}
\begin{figure}[!ht]
\centering
	\begin{minipage}[t]{0.49\columnwidth}
		\includegraphics[width=\linewidth]{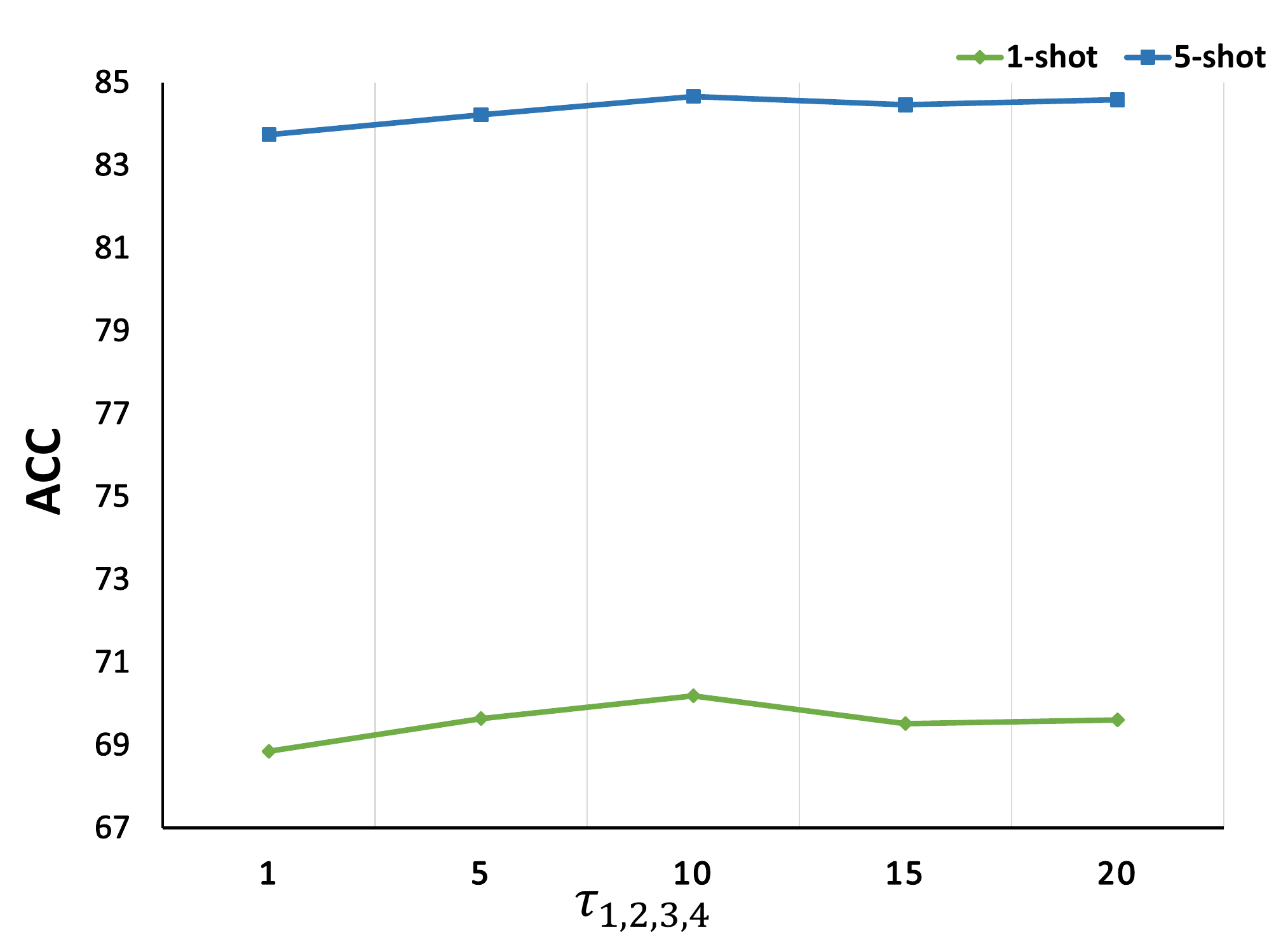}
        \subcaption{Fix $\alpha_{1,2,3}=1.0$}
        \label{fig:supp_pretrain_robust_a}
	\end{minipage}
	\begin{minipage}[t]{0.49\columnwidth}
		\includegraphics[width=\linewidth]{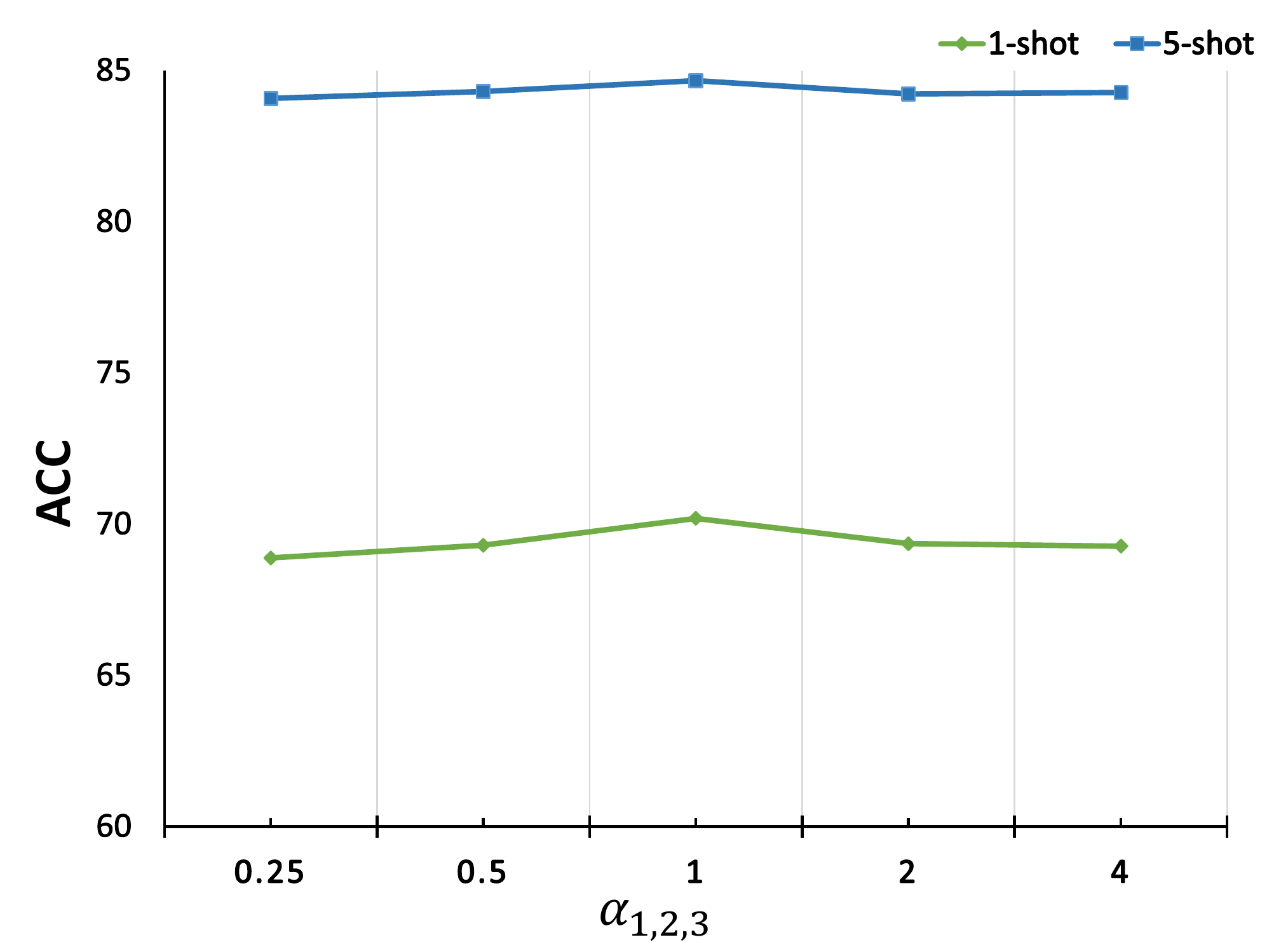}
        \subcaption{Fix $\tau_{1,2,3,4}=0.1$}
        \label{fig:supp_pretrain_robust_b}
	\end{minipage}
\caption{Effect of hyperparameters $\tau_{1,2,3,4}$ and $\alpha_{1,2,3}$ on miniImageNet.}
\label{fig:supp_pretrain_robust}
\end{figure}

We investigate the effect of hyperparameters in the pre-training stage on classification performance.
Note that we use the inverse temperature parameters in our implementation.
To make it easy to understand, we draw the graph according to the inverse temperature parameters.
That is, the horizontal axis of \figurename~\ref{fig:supp_pretrain_robust_a} actually represents the inverse of $\tau_{1,2,3,4}$.
As shown in \figurename~\ref{fig:supp_pretrain_robust},
we empirically find that the best results can be achieved by setting the four temperature parameters $\tau_{1,2,3,4}$ to 0.1 and the three balance scalars $\alpha_{1,2,3}$ to 1.0 at the same time.
The results indicate that our pre-training approach is less sensitive to the two kinds of hyperparameters (within a certain range), which means it is robust.
Thus it is effortless to apply our approach to other methods.

\newpage

\section{Quantitative Analysis of Model Parameters}
In the case of using the same backbone network, we compare the number of parameters of our proposed method with the two-stage methods ProtoNet~\cite{ProtoNet} and FEAT~\cite{FEAT}.
The results are shown in \tablename~\ref{tab:model_params}.
In the pre-training stage, the number of parameters of our method is 1.9M more than ProtoNet and baseline (FEAT)
due to the introduction of projection heads and a fully connected layer for our contrastive losses.
In the meta-training stage, a single projection head used in distance-scaled contrastive loss results in 0.4M more parameters for our method than baseline.
Furthermore, during meta-testing, the number of parameters of our method is the same as in baseline.
Therefore, the small number of additional parameters we introduce in the pre-training and meta-training stages is acceptable considering the improved classification performance of our method.

\begin{table}[!ht]
\centering
\caption{Comparison of the number of parameters for several methods.}
\renewcommand\arraystretch{1.2}
\begin{tabular}{l|l|cl|cl}
\hline
Stage                          & Method   & \multicolumn{2}{c|}{Backbone}                   & \multicolumn{2}{c}{\# params} \\ \hline
\multirow{3}{*}{Pre-training}  & ProtoNet~\cite{ProtoNet} & \multicolumn{2}{c|}{\multirow{9}{*}{ResNet-12}} & \multicolumn{2}{c}{12.5M}     \\
                               & FEAT~\cite{FEAT}     & \multicolumn{2}{c|}{}                           & \multicolumn{2}{c}{12.5M}     \\
                               & Ours     & \multicolumn{2}{c|}{}                           & \multicolumn{2}{c}{14.4M}     \\ \cline{1-2} \cline{5-6} 
\multirow{3}{*}{Meta-training} & ProtoNet~\cite{ProtoNet} & \multicolumn{2}{c|}{}                           & \multicolumn{2}{c}{12.4M}     \\
                               & FEAT~\cite{FEAT}     & \multicolumn{2}{c|}{}                           & \multicolumn{2}{c}{14.1M}     \\
                               & Ours     & \multicolumn{2}{c|}{}                           & \multicolumn{2}{c}{14.5M}     \\ \cline{1-2} \cline{5-6} 
\multirow{3}{*}{Meta-testing}  & ProtoNet~\cite{ProtoNet} & \multicolumn{2}{c|}{}                           & \multicolumn{2}{c}{12.4M}     \\
                               & FEAT~\cite{FEAT}     & \multicolumn{2}{c|}{}                           & \multicolumn{2}{c}{14.1M}     \\
                               & Ours     & \multicolumn{2}{c|}{}                           & \multicolumn{2}{c}{14.1M}     \\ \hline
\end{tabular}
\label{tab:model_params}
\end{table}

\newpage

\section{Qualitative Ablation Study}

\begin{figure}[!ht]
\centering
	\begin{minipage}[t]{0.49\columnwidth}
		\includegraphics[width=\linewidth]{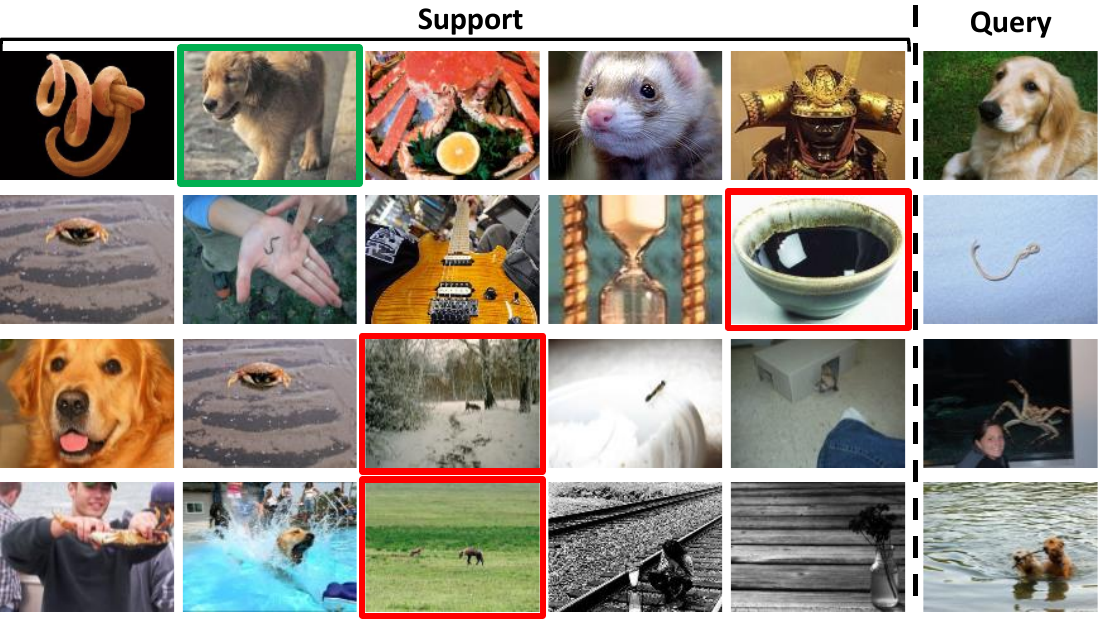}
        \subcaption{baseline}
        \label{fig:supp_ncm_baseline}
	\end{minipage}
	\begin{minipage}[t]{0.49\columnwidth}
		\includegraphics[width=\linewidth]{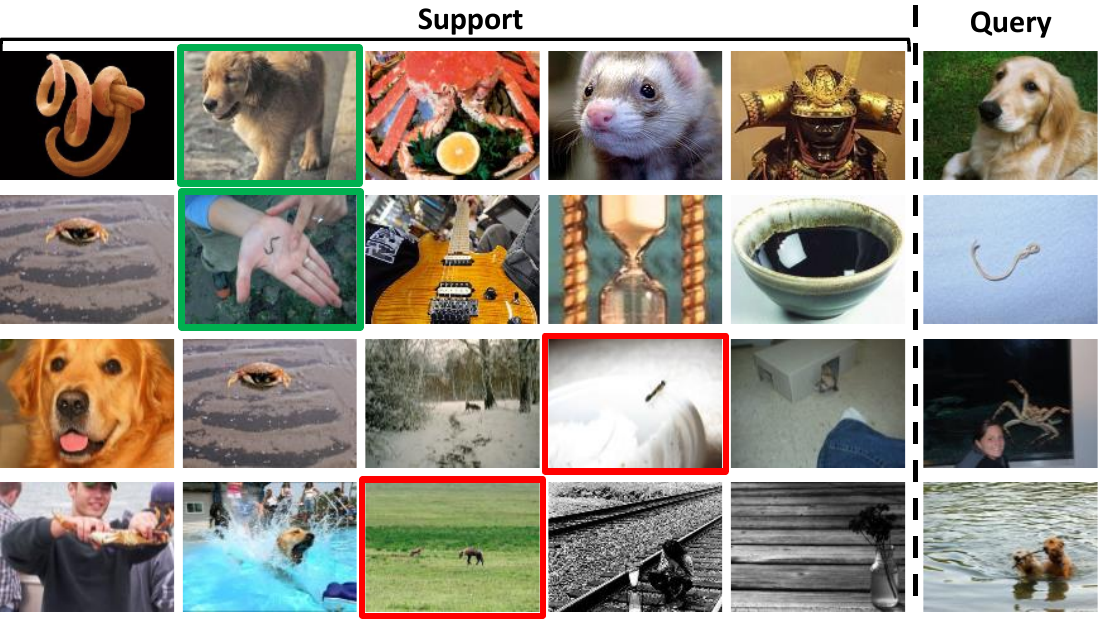}
        \subcaption{baseline+CL}
        \label{fig:supp_ncm_baseline+CL}
	\end{minipage}
    \\
	\begin{minipage}[t]{0.49\columnwidth}
		\includegraphics[width=\linewidth]{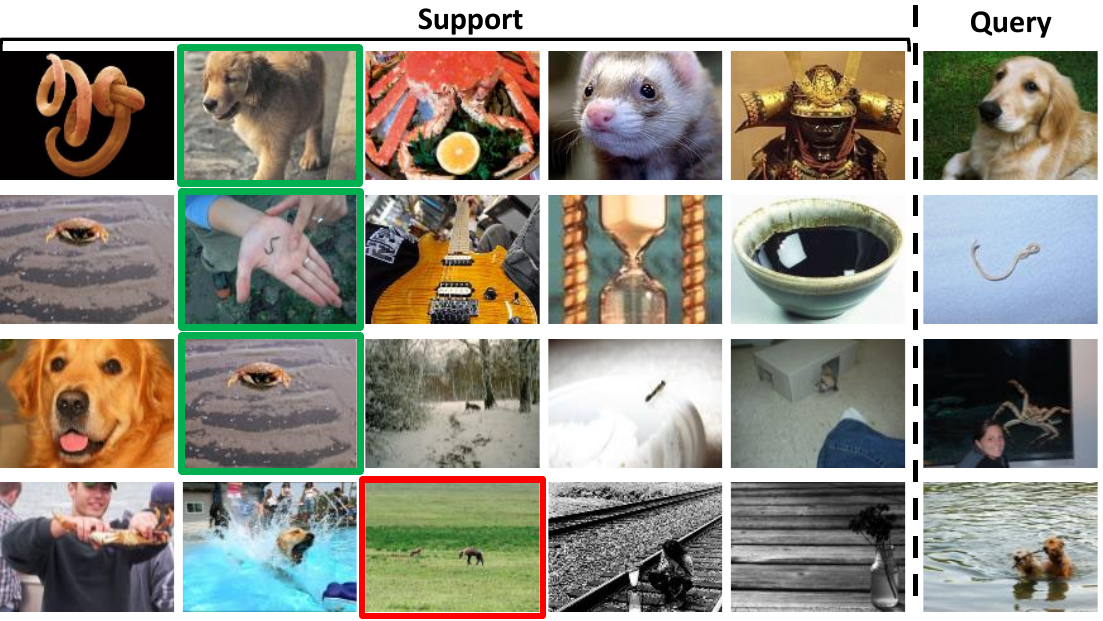}
        \subcaption{baseline+CL+CVET}
        \label{fig:supp_ncm_baseline+CL+CVET}
	\end{minipage}
	\begin{minipage}[t]{0.49\columnwidth}
		\includegraphics[width=\linewidth]{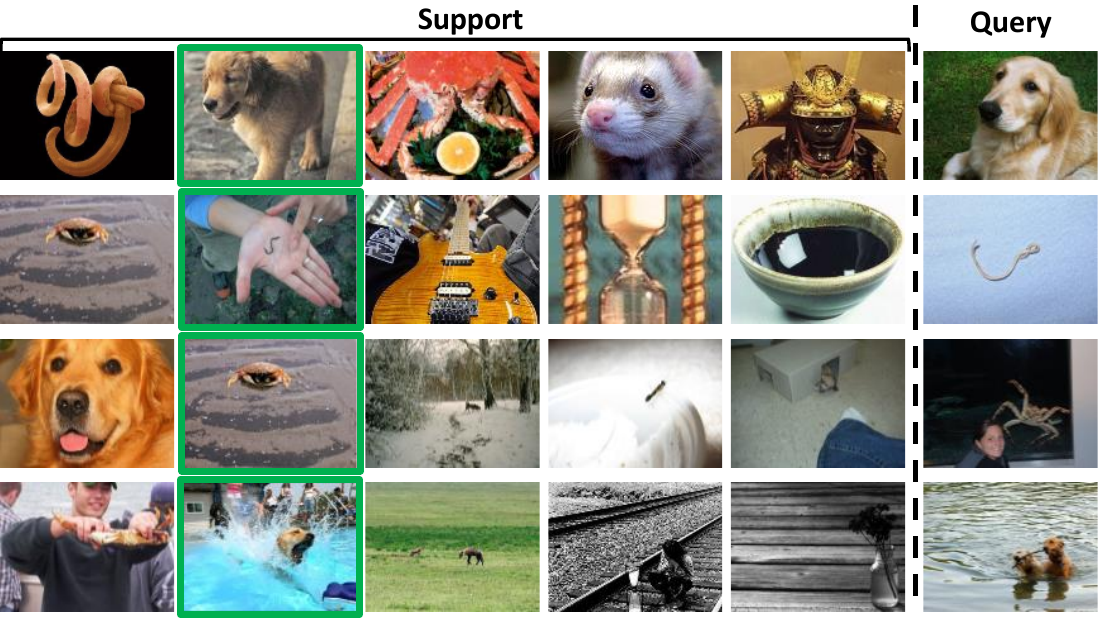}
        \subcaption{Ours}
        \label{fig:supp_ncm_ours}
	\end{minipage}
\caption{5-way 1-shot classification results on miniImageNet. The green box indicates the correct classification result, while the red box indicates the incorrect result.}
\label{fig:supp_ncm_results}
\end{figure}

In this section, we give qualitative analysis to verify the effectiveness of each component proposed in our method.
The classification results of our components are shown in \figurename~\ref{fig:supp_ncm_results},
Thanks to the successive use of each component, the model can better adapt to novel tasks with more pictures in which the background is dominant.
Moreover, the accurate recognition of small objects reflects that our framework enables the representations to learn meta-knowledge that is useful for few-shot classification.
Therefore, it can be considered that each of our proposed components is effective.

\newpage

\section{Visualization and Quantitative Analysis of Feature Embeddings}
\begin{figure}[!h]
    \centering
    \begin{minipage}[t]{0.32\columnwidth}
		\centering
        \includegraphics[width=1\linewidth]{ProtoNet.pdf}
        \subcaption{ProtoNet}
		\label{fig:supp_visual_protonet}
	\end{minipage}
	\begin{minipage}[t]{0.32\columnwidth}
		\centering
        \includegraphics[width=1\linewidth]{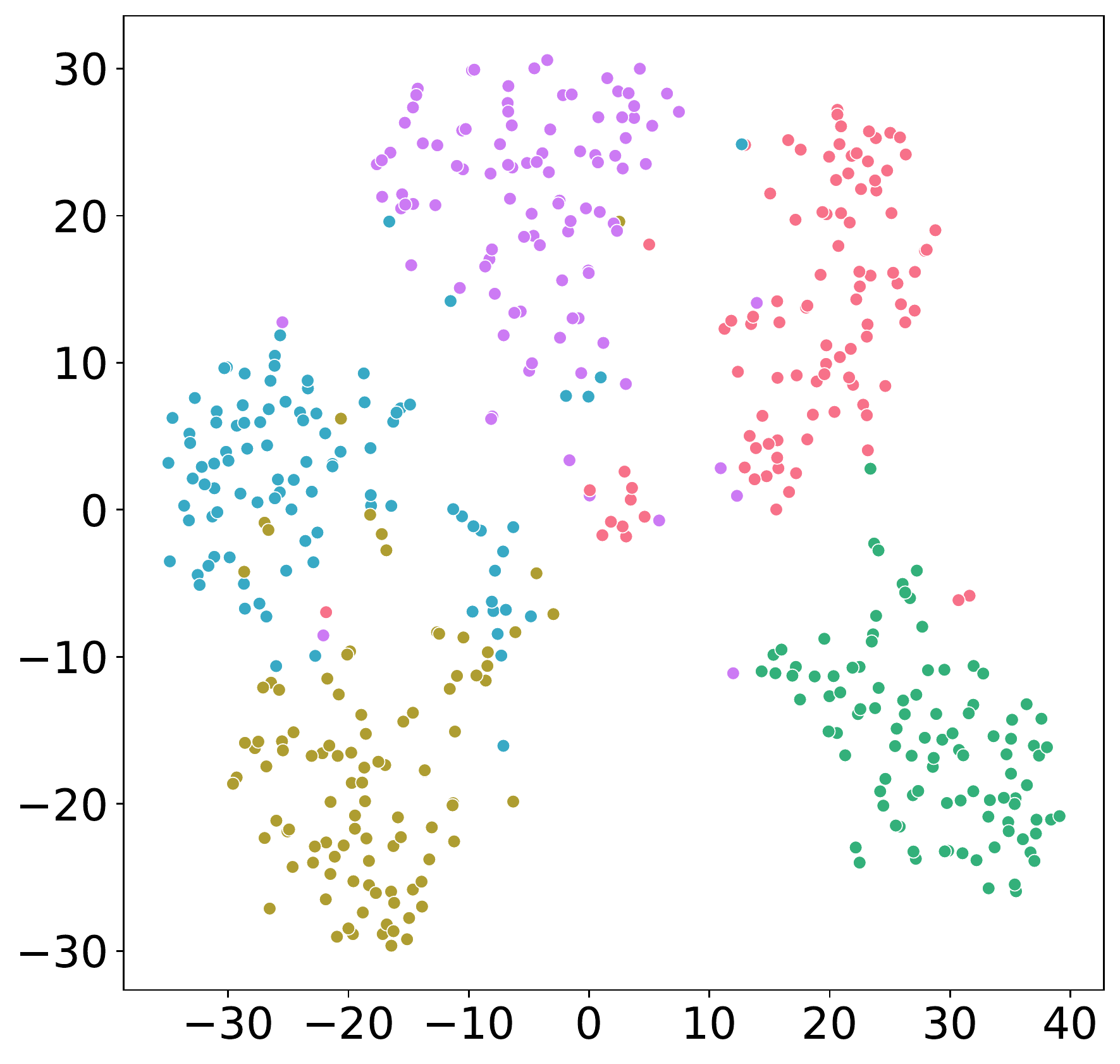}
        \subcaption{ProtoNet+CL}
		\label{fig:supp_visual_protonet_cl}
	\end{minipage}
	\begin{minipage}[t]{0.32\columnwidth}
		\centering
        \includegraphics[width=1\linewidth]{ProtoNet+Ours.pdf}
        \subcaption{ProtoNet+Ours}
		\label{fig:supp_visual_protonet_ours}
	\end{minipage}
    \\
    \begin{minipage}[t]{0.32\columnwidth}
		\centering
        \includegraphics[width=1\linewidth]{baseline.pdf}
        \subcaption{baseline}
		\label{fig:supp_visual_baseline}
	\end{minipage}
    \begin{minipage}[t]{0.32\columnwidth}
		\centering
        \includegraphics[width=1\linewidth]{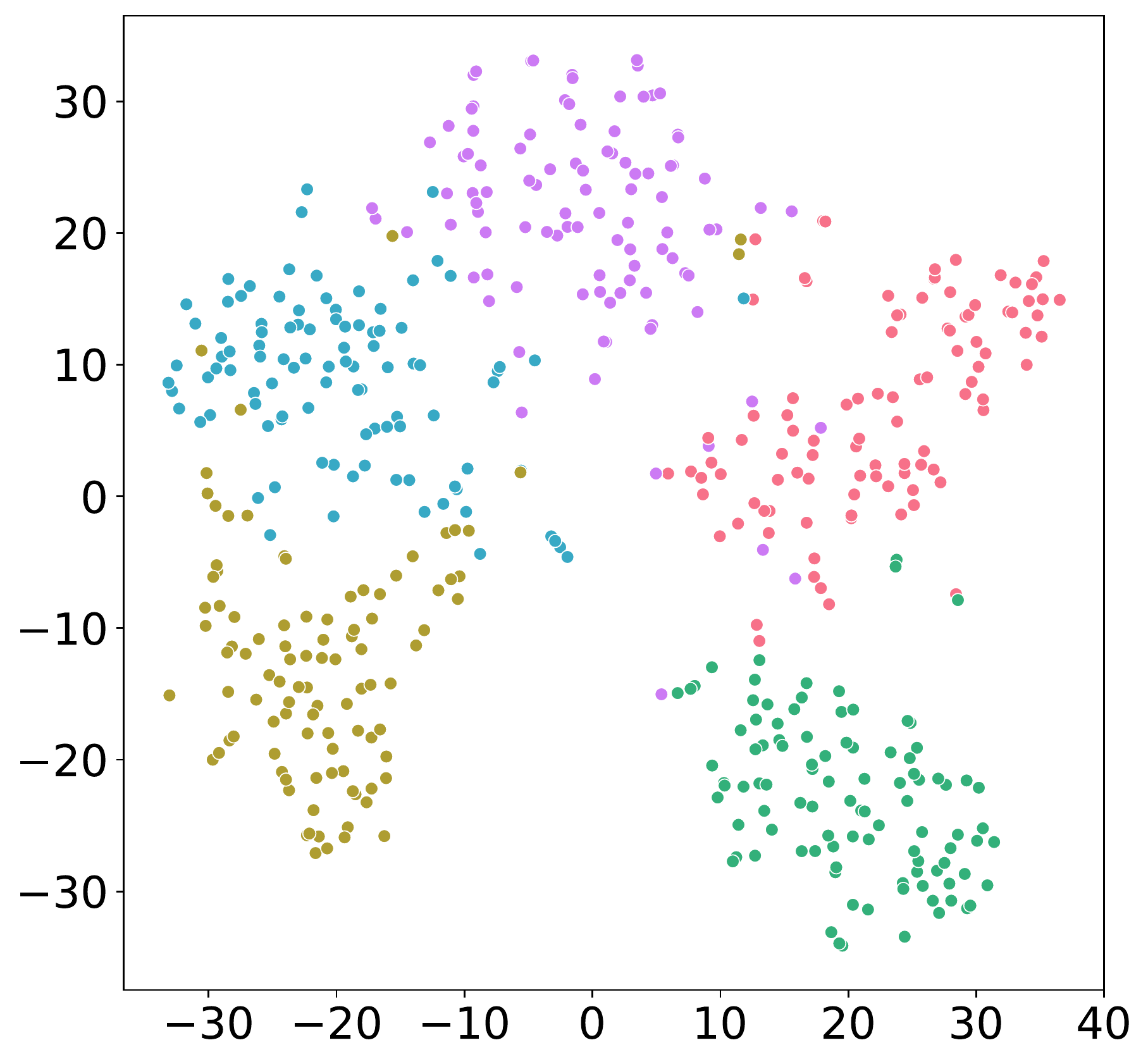}
        \subcaption{baseline+CL}
		\label{fig:supp_visual_baseline_cl}
	\end{minipage}
    \begin{minipage}[t]{0.32\columnwidth}
		\centering
        \includegraphics[width=1\linewidth]{Ours.pdf}
        \subcaption{Ours}
		\label{fig:supp_visual_ours}
	\end{minipage}
\caption{Visualization of 100 randomly sampled images for each of the 5 meta-test classes from miniImageNet by t-SNE.}
\label{fig:supp_visual_experiment}
\end{figure}

In this section, we present the complete visualization results of Sect.~4.4 in the paper.
The results from \figurename~\ref{fig:supp_visual_experiment} illustrate that our proposed framework generates more transferable and discriminative representations on novel classes.

\vspace{-10pt}

\vspace{-40pt}

\begin{table}[!ht]
\centering
\caption{Quantitative analysis of feature embeddings using SVM.}
\label{tab:svm_results}
\resizebox{0.4\columnwidth}{!}{%
\begin{tabular}{l|c}
\hline
Method        & Mean accuracy \\ \hline
ProtoNet      & 0.87          \\
ProtoNet+Ours & 0.91          \\
baseline      & 0.89          \\
Ours          & \textbf{0.98} \\ \hline
\end{tabular}%
}
\end{table}

We further give quantitative analysis below.
We use the same sampling strategy as in the visualization experiments above and divide the training and test sets in a 1:4 ratio.
Then the features extracted by the four methods ProtoNet, ProtoNet+Ours, baseline and Ours are classified with SVM.
The mean accuracies of these methods are shown in \tablename~\ref{tab:svm_results}.
The results indicate that the two-stage methods combined with our framework make feature embeddings more separable.

\end{document}